\documentclass[sigconf, 10pt,nonacm]{acmart}
\acmConference{}{}{}
\usepackage{array}
\usepackage{subcaption}
\usepackage{amsmath}

\usepackage{soul}
\AtBeginDocument{%
  }

\setcopyright{none}
\renewcommand\footnotetextcopyrightpermission[1]{}
\newcommand{\myparagraph}[1]{{\noindent \bf #1}}

\newcommand{\resnet}{ResNet-26}
\newcommand{\resnetF}{ResNet-50}
\newcommand{\vit}{ViT}

\newcommand{\cifar}{CIFAR-10}
\newcommand{\cifarc}{CIFAR-10C}
\newcommand{\pacs}{PACS}
\begin{document}

\title{BoTTA: Benchmarking on-device Test Time Adaptation}
% for arXiv
\author{Michal Danilowski}
\affiliation{
  \institution{University of Birmingham}
  \city{Birmingham}
  \country{United Kingdom}
}
\email{mxd411@student.bham.ac.uk}

\author{Soumyajit Chatterjee}
\affiliation{
  \institution{Nokia Bell Labs}
  \city{Cambridge}
  \country{United Kingdom}
}
\email{sjituit@gmail.com}

\author{Abhirup Ghosh}
\affiliation{
  \institution{University of Birmingham}
  \city{Birmingham}
  \country{United Kingdom}
}
\email{a.ghosh.1@bham.ac.uk}

\begin{abstract}
The performance of deep learning models depends heavily on test samples at runtime, and shifts from the training data distribution can significantly reduce accuracy. Test-time adaptation (TTA) addresses this by adapting models during inference without requiring labeled test data or access to the original training set. While research has explored TTA from various perspectives like algorithmic complexity, data and class distribution shifts, model architectures, and offline versus continuous learning, constraints specific to mobile and edge devices remain underexplored.

We propose BoTTA, a benchmark designed to evaluate TTA methods under practical constraints on mobile and edge devices. Our evaluation targets four key challenges caused by limited resources and usage conditions: (i) limited test samples, (ii) limited exposure to categories, (iii) diverse distribution shifts, and (iv) overlapping shifts within a sample. We assess state-of-the-art TTA methods under these scenarios using benchmark datasets and report system-level metrics on a real testbed. Furthermore, unlike prior work, we align with on-device requirements by advocating periodic adaptation instead of continuous inference-time adaptation.

Experiments reveal key insights: many recent TTA algorithms struggle with small datasets, fail to generalize to unseen categories and depend on the diversity and complexity of distribution shifts. BoTTA also reports device-specific resource use. For example, while SHOT improves accuracy by $2.25\times$ with $512$ adaptation samples, it uses $1.08\times$ peak memory on Raspberry Pi versus the base model. BoTTA offers actionable guidance for TTA in real-world, resource-constrained deployments.
\end{abstract}

\maketitle
\section{Introduction}
Due to distribution shifts between training and deployment environments, deep learning models often suffer from performance degradation when deployed in real-world settings. This problem is particularly pronounced in mobile and edge devices, which operate across diverse spatiotemporal contexts, capturing data under varying environmental conditions, user behaviors, and hardware constraints. Unlike centralized systems, where deployment settings can be controlled, mobile devices continuously encounter unseen domains, making it infeasible to pre-train models for all scenarios. As a result, models trained on a source domain often fail to generalize effectively to the evolving target domain.

To mitigate this, different domain adaptation strategies refine model parameters using real-world test data (called target domain) to improve the model performance in the deployment environment. Initial efforts explored transfer learning requiring labeled target data~\cite{pmlr-v32-donahue14}. However, as annotating target data is challenging, especially in mobile and edge devices, unsupervised domain adaptation (UDA)~\cite{sun2019unsupervised, ganin2016domain, wilson2020survey, baktashmotlagh2013unsupervised,10.1145/3380985} enabled adaptation on unlabeled target data. Still, UDA require source training data which can be impractical considering privacy, communication costs, and constrained storage space on mobile devices.

Recently proposed Test Time Adaptation (TTA) methods addressed both limitations by adapting on unlabeled target data without accessing training set~\cite{wang2020tent, niu2023towards, wang2022continual, liang2020we, gong2023note,gong2023sotta,chen2022contrastive}. Popular families of TTA algorithms include minimizing the entropy of the class probability distribution~\cite{wang2020tent}, optimizing pseudo-labels~\cite{liang2020we}, using contrastive learning~\cite{chen2022contrastive}, and tuning the classifier without optimization~\cite{iwasawa2021test}. Given their unsupervised approach, TTA techniques are particularly well-suited for on-device settings, where data annotation is hard and source datasets may be inaccessible due to privacy or storage constraints.

\begin{figure}[]
    \centering
    \includegraphics[width=\linewidth,keepaspectratio]{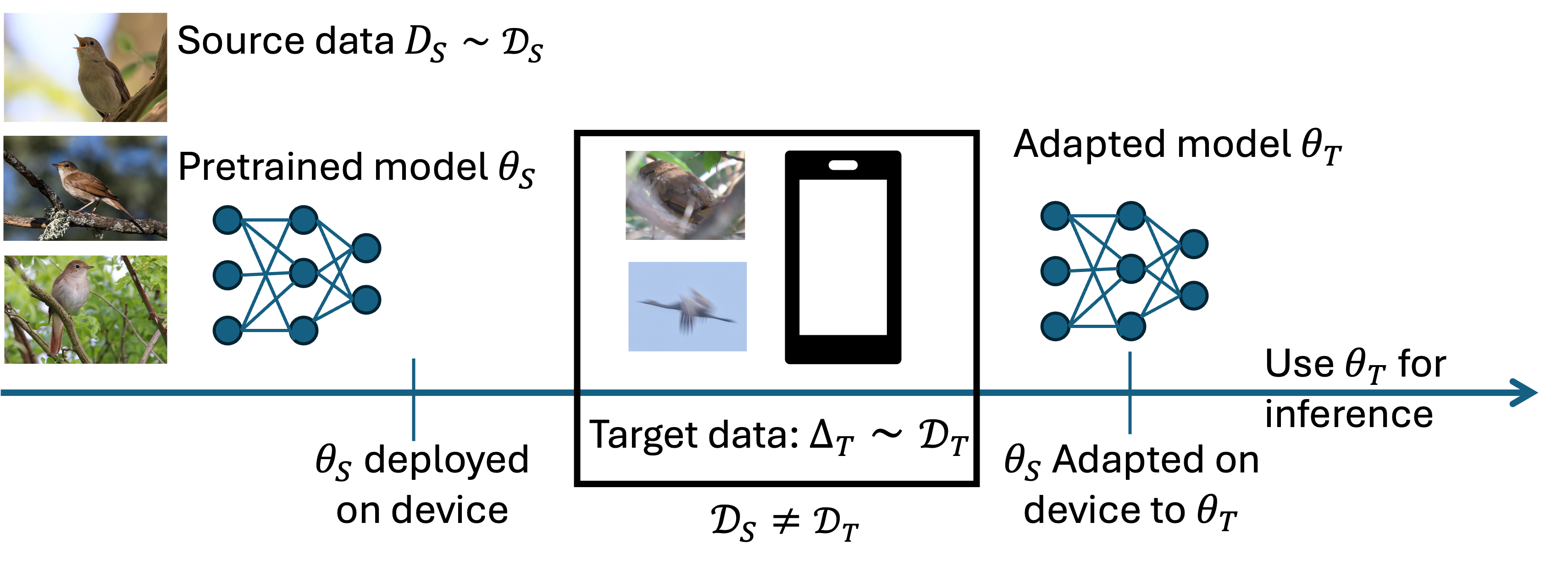}
    \caption{On-device domain model adaptation settings with an example bird classifier. The pre-trained model (on clean images) is deployed on a phone. The pictures of the birds that the user takes are noisy, e.g., blurred. The model is adapted using the target data and remains frozen for inference thereafter.}
    \label{fig:setting}
\end{figure}

Despite their potential for deployment on mobile and edge devices (\figurename~\ref{fig:setting}), many existing TTA approaches are not explicitly designed considering the constraints presented by mobile devices. For example, a common assumption of many TTA approaches is the availability of large batches of target data~\cite{wang2020tent,liang2020we}, which can be infeasible on resource-constrained devices with limited memory and usage constraints. Some methods further exacerbate this issue by requiring separate memory banks or multiple copies of the model’s encoder~\cite{chen2022contrastive}. Additionally, the standard TTA evaluation protocol assumes that adaptation and inference occur simultaneously, meaning that models are continuously updated at test time while making predictions~\cite{wang2020tent,gong2023note}. However, in real-world edge deployments, adaptation opportunities may be sporadic, driven by the availability of resources or user interactions, rather than occurring alongside every inference step. This gap between existing TTA design choices and the realities of mobile deployment scenarios motivates the need for a tailored benchmarking framework that can better reflect real-world constraints and assess the key limitations of TTA algorithms considering these constraints.

Notably, many recent works have attempted to benchmark TTA approaches by considering factors such as algorithmic complexity, domain and class distributions, and impact of model architectures~\cite{zhao2023pitfalls, NEURIPS2023_7d640f37,yu2023benchmarking,alfarra2023revisiting,du2024unittaunifiedbenchmarkversatile,fgk}. These benchmarks have successfully identified key limitations, including high hyperparameter sensitivity, strong dependence on the source model architecture, and limited robustness to shifts in data and class distributions. However, none of these studies systematically evaluate how TTA methods perform under the constraints imposed by mobile and edge devices. 

\myparagraph{This work.} To bridge the above gap we introduce BoTTA to benchmark state-of-the-art TTA methods in scenarios that explicitly account for the challenges of real-world edge deployments. We design a diverse set of evaluation conditions that reflect realistic constraints, including adaptation with a limited number of target samples, adaptation with incomplete class coverage, varying severity and types of domain shifts, and overlapping corruptions within individual samples. Additionally, we depart from the standard practice of evaluating TTA models during continuous adaptation and instead assess their effectiveness under periodic adaptation, a more realistic setting for mobile devices where adaptation opportunities are intermittent. Through extensive experiments on \cifarc{}~\cite{hendrycks2019benchmarking} and \pacs{}~\cite{li2017deeper} datasets, evaluated on resource-constrained devices (Raspberry Pi 4B and Jetson Orin Nano), we provide new insights into the strengths and limitations of current TTA techniques when applied in real-world edge settings. Our key contributions are as follows.
\begin{enumerate}
    \item BoTTA benchmarks data availability constraints under realistic scenarios: $(i)$ limited target samples, $(ii)$ incomplete class coverage during adaptation, $(iii)$ diverse distribution shifts to adapt, and $(iv)$ overlapping distribution shifts in individual samples, highlighting challenges not previously studied in TTA literature.
    \item Unlike conventional TTA studies that assume simultaneous adaptation and inference, aligned with resource constrained edge devices BoTTA evaluates models under periodic adaptation, reflecting real-world constraints where adaptation opportunities are sporadic and must be efficiently utilized.
    \item BoTTA presents a comprehensive set of experiments using two datasets (\cifarc{} and \pacs), and three model architectures (\resnet, \resnetF, and \vit). It produces a suit of interesting observations along with their technical rationale.
    \item We assess computational feasibility by benchmarking state-of-the-art TTA approaches on real edge devices (Raspberry Pi 4B, Jetson Orin Nano), measuring memory usage, CPU load, and adaptation overhead to provide actionable insights for practical deployment.
\end{enumerate}
We first motivate the case of on-device model adaptation in real world applications (Section~\ref{sec:background}) followed by a mathematical formulation of the problem solved by TTA (Section~\ref{sec:formal-tta}). Next, a brief survey of the TTA benchmarking efforts in Section~\ref{sec:rel_work} places BoTTA in the appropriate research landscape highlighting its novelty. Then we introduce the benchmarking framework in Section~\ref{sec:settings} with detailed description of the scenarios and their evaluation strategies. The rest of the paper (Section~\ref{sec:scenario1} -- \ref{sec:scenario4}) presents observations on the proposed scenarios including the on-device system performance evaluation in Section~\ref{sec:on_device}. Finally, we consolidate with a list of key takeaways in Section~\ref{sec:keytake} and conclude in Section~\ref{sec:conclusions}.

\section{Motivation and Background}
This section motivates the necessity of on-device domain adaptation followed by a brief overview of the algorithmic studies in TTA methods, and a formal setting for TTA. Readers familiar with TTA may skip Section~\ref{sec:prior_work_TTA}. 
\label{sec:background}
\subsection{Motivating On-device Model Adaptation}
As edge devices are used in critical applications, performance drops due to target domain shifts can have serious health, economic, and societal consequences. For example, pulse oximeters may overestimate blood oxygen levels in people with darker skin, leading to racially biased delayed care during COVID-19~\cite{bbcCovidPulse,bbcMedicalDevice}. In Indonesia's 2024 presidential election, an ML-based digit classification was used to infer the vote counts from mobile phone photos in every polling booths around the country. These counts were then aggregated centrally. However, the digit recognition failed on many occasions due to domain shifts (e.g., camera quality, lighting) resulting in wrong vote counts causing political turmoil in the country~\cite{jakartaglobeTechnicalIssues}. Similarly, speech recognition used for auto-populating patient records in GP practices~\cite{heidihealthHeidiHealth} risks failure due to accent variations, environmental conditions, microphone quality, or lack of cultural knowledge.

Traditionally, to address domain shift, model parameters are tuned in a server given a large target dataset gathered from multiple edge devices, and later the modified model is deployed through over-air-channel to the edge devices. However, the approach has the following drawbacks: $a)$ due to the inherent delay in the above process, the on-device model does not always remain accurate. As the inference from on-device models is fed directly to the user or other systems, such inaccuracy has a high cost. $b)$ Sensitive data from the edge devices needs to flow to possibly untrusted servers, raising privacy concerns.

Therefore, in this paper, we argue that the models need to be adapted on edge devices. While this theoretically is possible due to the availability of data and compute power at the edge devices, there are several challenges in this process. Throughout the paper, we consider two types of applications involving a) personal devices like mobile phones, and smartwatches, and b) fixed infrastructure monitoring devices such as security cameras. The on-device model adaptation is significantly influenced by the unique data and resource characteristics, which is the primary focus of our paper.

\subsection{Test Time Adaptation: A Practical Solution to Distribution Shifts}\label{sec:prior_work_TTA}
To address domain shift at test time, transfer learning~\cite{pmlr-v32-donahue14} is a common approach but is limited by the lack of labeled target data. Unsupervised domain adaptation (UDA)~\cite{sun2019unsupervised, ganin2016domain, wilson2020survey, baktashmotlagh2013unsupervised,10.1145/3380985} leverages unlabeled target data but requires computationally intensive adversarial training and access to labeled source data which might be challenging to obtain for privacy-sensitive applications. Another alternative presented in Test-time training (TTT)~\cite{sun19ttt,gandelsman2022testtime,10.1145/3638550.3643618} adapts models using a multitasking architecture and self-supervised auxiliary tasks but assumes control over the source regime and the model architecture, which may not always be feasible~\cite{chen2022contrastive}.

Notably, test-time adaptation (TTA) offers a more flexible approach by adapting the models at test time without requiring modifications to the model architecture or access to the source training dataset~\cite{wang2020tent, niu2023towards, wang2022continual, liang2020we, gong2023note, gong2023sotta, chen2022contrastive}. As a proxy for the labels at the runtime, \emph{softmax entropy} is often used as the optimization objective for adapting the model~\cite{wang2020tent}. However, one of the key limitations of this approach is that it can degrade the overall calibration of the model, making it more certain even for wrong predictions~\cite{chen2022contrastive}. Alternate approaches like SAR~\cite{niu2023towards} and SoTTA~\cite{gong2023sotta} extend this by minimizing the softmax entropy while keeping the optimization process \emph{sharpness-aware}. Other approaches include works like SHOT~\emph{pseudo-labeling}-based approach~\cite{liang2020we}. Although all these works take different approaches to performing TTA, they all fine-tune the encoder or specific layers of the encoder, like the batch-norm or group-norm layers, to adjust the model. Interestingly, works like T3A~\cite{iwasawa2021test} and OFTTA~\cite{10.1145/3631450} consider a completely different technique by taking a class prototype-based optimization-free approach where, unlike the previous works, they do not fine-tune the encoder and instead focus on aligning the classifier with the domain-shifted data.
\begin{table}[]
    \centering
    \caption{Important symbol definitions.}
    \label{tab:symbols}
    \begin{tabular}{c|c}
         Symbol & Description\\ 
         \hline
         $D_T$ & Full target domain data in the universe \\
         $\Delta_T$ & On-device adaptation dataset ($\Delta_T \subseteq D_T$) \\
         $\theta_S$ & Pre-trained model on source data, $D_S$\\
         $\theta_m$ & Model adapted on the dataset $m$\\
         $\xi(.)$ & Evaluation metric, e.g., accuracy\\
         $\mathcal{C}(D)$ & Set of sample categories in dataset $D$\\
          $\mathcal{S}(D)$ & Set of domains in dataset $D$\\
         \hline
    \end{tabular}
\end{table}

One key aspect of all these aforementioned approaches is that they do not consider the on-device efficiency of adaptation as the primary objective and instead focus only on the accuracy of adaptation. Recently, some of the works in TTA have started looking into designing efficient TTA approaches by considering smaller batch sizes~\cite{10.24963/ijcai.2024/616}, class-imbalance~\cite{su2023realworldtesttimeadaptationtrinet}, catastrophic forgetting~\cite{niu2022efficient}, memory-efficiency~\cite{song2023ecotta}, for micro-controller-based architectures~\cite{jia2024tinytta}, for non-parametric classifiers~\cite{zhang2023adanpc}, and model robustness~\cite{huang2023fourier}. Although these works discuss the efficiency of TTA approaches on-device, they usually consider different model architectures or setups that deviate from classical TTA setups. For example, TinyTTA~\cite{jia2024tinytta} assumes early exit architecture and does not consider general DNNs for adaptation. In BoTTA, we do not benchmark these algorithms as they require additional assumptions that may be unrealistic for edge and mobile devices.

\subsection{The Typical Problem Solved using TTA}\label{sec:formal-tta}
A classification model, $\theta_S$ trained on a source training dataset, $D_S = \{X_S^{(i)}, y_S^{(i)})\}$ is typically deployed at the edge devices for on-device inference. In the wild, the model is then tasked with the prediction on a target dataset $D_T = \{X_T^{(i)}\}$ where the distribution $X_T^{(i)} \sim \mathcal{X}_T$ differs from $X_S^{(i)} \sim \mathcal{X}_S$, commonly known as covariate shift. While this inherently changes the joint distribution of $\mathcal{D}_S \sim (X_S^{(i)}, y_S^{(i)})$, in this paper, we only consider the shifts in the input data distributions in line with the typical TTA literature~\cite{wang2020tent}. In this paper, we only focus on classification problems where the set of classes remains the same across the source and target data, $\mathcal{C}(D_S) = \mathcal{C}(D_T)$ where $\mathcal{C}(D)$ denotes the set of classes in the dataset $D$.

Let's assume $\mathcal{S}(D)$ denotes the set of domains in the dataset $D$. Further, one can model a shift as a hypothetical function $\chi: X_S^{(i)} \times \tau \times \mu \rightarrow X_T^{(i)}$ transforming a clean data sample, $X_S^{(i)}$ according to a corruption type $\tau$ and severity $\mu$. In practice, such $\chi$ is unknown and does not have a closed form, which is the primary reason behind the model adaptation being a complex problem. 

Notice that the target dataset, $D_T$ is unlabeled, making the case of Test Time Adaptation. Therefore, a TTA method is the following mapping, $\mathcal{A} : \theta_S \times D_T \rightarrow \theta_{D_T}$.

Consider an evaluation metric, $\xi$ to evaluate the performance of a model, $\theta$ on a dataset $D$ defined as a mapping $\xi: \theta \times D \rightarrow \mathbb{R}$, e.g., accuracy. The objective of any TTA method is to produce a model, $\theta_{D_T}$ such that $\xi(\theta_{D_T}, D_T) > \xi(\theta_S, D_T)$. Typically, an accuracy-related metric is of interest, however, such $\xi$ is impossible to empirically derive due to the unavailability of ground truth labels, $y_T$. Therefore, TTA methods often optimize a proxy, for example, minimize the entropy of the class distributions, $\mathcal{L}_{ent}$ on target samples $X_T$. These methods use popular gradient descend algorithms to update a selected part of the neural network to find the minimizer to the objective functions of the following form  $\theta_T = argmin_{\theta} \mathcal{L}_{ent}(X_T)$.
\section{Prior Benchmarking of TTA Algorithms}
\label{sec:rel_work}

Recent studies have identified critical challenges in TTA and proposed new benchmarking strategies to assess its feasibility in real-world settings. TTAB~\cite{zhao2023pitfalls} systematically evaluates ten state-of-the-art TTA methods across diverse distribution shifts and reveals three key limitations: hyperparameter sensitivity, model dependency, and limited robustness. 

Continuously Changing Corruptions~\cite{NEURIPS2023_7d640f37} benchmarks adaptation stability over extended periods. Computational efficiency has also been benchmarked~\cite{10.5555/3692070.3692111, alfarra2023revisiting} showing AdaBN and SHOT outperform more complex methods in real-time adaptation scenarios. Further,~\cite{yu2023benchmarking} compares different genres of TTA namely test-time batch adaptation (TTBA), online test-time adaptation (OTTA), and test-time domain adaptation (TTDA). UniTTA~\cite{du2024unittaunifiedbenchmarkversatile} uses a Markov state transition framework to simulate $36$ scenarios including i.i.d., non-i.i.d., continual, and imbalanced data distributions. Another similar benchmarking study explored in~\cite{liang2023ttasurvey} focuses on performance of TTA approaches under significant distribution shifts.
Further,~\cite{fgk} evaluates prompt-based techniques for adapting Vision Language Models.

Collectively, these studies emphasize the need for TTA methods that balance adaptation performance, computational efficiency, and long-term robustness to ensure practical deployment on resource-constrained devices. From above benchmarking studies, the following are the key gaps.

\begin{enumerate}
    \item Most existing benchmarking approaches focus on the model centric performance of TTA and miss data centric impacts, i.e., they lack analysis of multiple corruption types and adaptation under limited categories.
    \item Existing TTA benchmarks do not evaluate TTA algorithms considering practical constraints like limitations of samples or diversity in distribution shifts which are prevalent in edge-based applications.
    \item Prior TTA benchmarks do not sufficiently evaluate the algorithms on real-life testbeds that simulate edge-devices' resource constraints.
\end{enumerate}
Recognizing the limitations in existing benchmarking strategies, this paper introduces BoTTA with a set of tailored scenarios designed to evaluate the feasibility of deploying current TTA algorithms on edge and mobile devices. The details of these scenarios are as follows.
\section{Benchmarking Settings for BoTTA}
\label{sec:settings}
This section defines our benchmarking scenarios along with their rationale corresponding to edge device applications followed by the data and model settings for the evaluation.

\subsection{Key Benchmarking Scenarios}
Each scenario in the following targets a specific aspect of the domain adaptation problem. They all inspect the behavior of the TTA methods given an adaptation dataset, $\Delta_T \subseteq D_T$. The scenarios differ in how $\Delta_T$ is selected: with varying $|\Delta_T|$, categories available in $\Delta_T$, diversity of the shifts in $\Delta_T$, and complexity of the distribution shifts.

To benchmark TTA methods, in line with a typical edge application requirement, we consider the evaluation metric, $\xi(\theta_{\Delta_T}, D)$ to be the accuracy with a varying definition of $D$. While the accuracy is impossible to measure in practice as $y_T$ is unavailable, in line with the TTA literature, we assume its access in our performance reporting tool (while the adaptation tool can't access it). Further, our evaluation method departs from the standard practice in TTA algorithms (Section~\ref{sec:model-eval-strategy}). More implementation details on the scenarios are discussed along with their observations in Section~\ref{sec:scenario1}-- Section~\ref{sec:scenario4}.

\subsubsection{Scenario $1$: Limited target samples}\label{sec:scenario-def-1}
A critical (yet overlooked) assumption in all the algorithmic research on TTA is to have access to a large target dataset for adaptation. Below we argue it is unreasonable in the edge device settings and correspondingly define an edge-friendly evaluation strategy.

\myparagraph{Rationale.}
The size of the target data, $\Delta_T$ available for adaptation is limited due to two properties in edge applications.
\begin{itemize}
    \item Edge applications either produce data with active user participation or as a result of automated sensing. Former types of applications naturally take a long time to record a large target dataset. And certain applications in the latter category naturally produce data at a slow rate, for example, a sleep-tracking application produces useful data chunks every night, and thus $|\Delta_T|$ scales linearly with the number of nights observed.

    \item Even if an application generates large datasets at a faster rate, as edge devices have limited storage, a limited number of target samples will be available for adaptation.
\end{itemize}

Note that this consideration differs from the literature in continual Test Time Adaptation~\cite{wang2022continual} and the way typical TTA methods~\cite{wang2020tent} evaluate the adapted models: evaluate and adapt with the same batch of samples. They both assume a large dataset is available either through a streaming or random access. In contrast, we argue edge devices will have random access to a small target dataset.

Motivated by the above, we define our following scenario.

\myparagraph{Definition.} Adapt a pretrained model, $\theta_S$ on a target dataset $\Delta_T \subseteq D_T$ using different TTA methods, $\mathcal{A}: \theta_S \times \Delta_T \rightarrow \theta_{\Delta_T}$ and consequently report the performance of $\theta_{\Delta_T}$ using accuracy on $\Delta_T$ and $D_T$.

As $|\Delta_T|$ is application and device-dependent, we benchmark using multiple values. We measure accuracy based on two test sets, $\Delta_T$ and $D_T$ where the former represents the typical evaluation strategy in TTA literature and the latter presents the typical way to test model generalization through a large test set. This follows from the fact that both $\Delta_T$ and $D_T$ are drawn from the same data distribution, $\mathcal{D}_T$.

\subsubsection{Scenario $2$: Limited categories to adapt}\label{sec:S2-defn}
An important assumption in all TTA algorithms is that the target data contains all the categories as in the source data, i.e., $\mathcal{C}(D_S) = \mathcal{C}(\Delta_T)$.

\myparagraph{Rationale.} The above assumption does not align edge device settings as a device user typically has limited exposure to the world, restricting the categories in $\Delta_T$. For example, in a task to recognize natural objects, $\mathcal{C}(\Delta_T)$ will be limited by the user's geography. Thus, a reasonable application assumption is that a user adapts their model to a dataset that contains a subset of the categories to achieve a generalized model. Technically, as the covariate shift is assumed to affect all the data categories similarly (e.g., every type of outdoor image is affected similarly when it is foggy), a reasonable expectation is that $\theta_{\Delta_T}$ will be accurate on all categories irrespective of whether it is used for adaptation.

\myparagraph{Definition.} Adapt a source model, $\theta_S$ on a target dataset $\Delta_T \subseteq D_T$ such that $\mathcal{C}(\Delta_T) \subseteq \mathcal{C}(D_T)$ using different TTA methods, $\mathcal{A}: \theta_S \times \Delta_T \rightarrow \theta_{\Delta_T}$ and consequently report the performance of $\theta_{\Delta_T}$ using accuracy on $\Delta_T$, $D_T$, and $\Delta_{OOD}$.

Here $\Delta_{OOD}$ denotes the out-of-distribution data in terms of the categories included, formally, $\mathcal{C}(\Delta_{OOD}) \cap \mathcal{C}(\Delta_T) = \phi$ where $\phi$ denotes the empty set. $\mathcal{C}(\Delta_T)$ is application dependent, thus we test with different combinations.

\subsubsection{Scenario $3$: Diverse distribution shifts in $\Delta_T$}\label{sec:S3-defn}
An assumption in all TTA methods to simplify the adaptation problem is to consider that all samples in $\Delta_T$ suffer from the same type of domain shift.

\myparagraph{Rationale.} However, it is an unreasonable assumption for edge devices. For example, outdoor photos will have various types of shifts such as snow, rain, and fog depending on the weather. Thus, here we design a novel evaluation scenario where $\Delta_T$ contains data samples from various distribution shifts.

\myparagraph{Definition.} Given a set of domains $\mathcal{S}_j$ with varying types and amounts of distribution shifts, a device contains an adaptation dataset with $\mathcal{S}(\Delta_T) \subseteq \mathcal{S}(D_T)$ domains. Adapt a pretrained model, $\theta_S$ on $\Delta_T$ using different TTA methods, $\mathcal{A}: \theta_S \times \Delta_T \rightarrow \theta_{\Delta_T}$ and consequently report the performance of $\theta_{\Delta_T}$ using accuracy on three test sets: $D_T$, $\Delta_T$, and $\Delta_{OOD}$.

Here $\Delta_{OOD}$ denotes the out-of-distribution data in terms of the domains included, formally, $\mathcal{S}(\Delta_{OOD}) \cap \mathcal{S}(\Delta_T) = \phi$. $\mathcal{S}(\Delta_T)$ is application dependent, thus we test with different combinations.

\subsubsection{Scenario $4$: Overlapping distribution shifts in $\Delta_T$} \label{sec:S4-defn} Another simplifying assumption in all TTA methods is to consider the same type and severity of domain shift in $\Delta_T$. However, this departs from a typical edge device application scenario.

\myparagraph{Rationale.} In edge device settings, a single data sample may suffer from a overlapping set of corruptions, for example, an outdoor photo may be taken in foggy weather with motion blur. Therefore, in this scenario, we consider that a data sample may suffer from multiple corruptions.

\myparagraph{Definition.} Given a set of types of corruption, $\{\tau_j\}$ and severity, $\{\mu_j\}$, a device has an adaptation dataset as $\Delta_T = \chi(\cdots\chi(\chi(D_S, \tau_0, \mu_0), \tau_1, \mu_1), \cdots, \tau_{n}, \mu_{n})$. Adapt a pretrained model, $\theta_S$ on $\Delta_T$ using different TTA methods, $\mathcal{A}: \theta_S \times \Delta_T \rightarrow \theta_{\Delta_T}$ and consequently report the performance of $\theta_{\Delta_T}$ using accuracy on $D_T$, containing a larger number of samples as $\Delta_T$ but following the same distributions, i.e., the same combinations of the corruptions.

\subsubsection{Scenario 5: On-device system resource evaluation}\label{sec:S5-defn}
 
Unlike cloud-based adaptation, edge devices have limited CPU and memory capacity, which can impact the feasibility and efficiency of on-device test-time adaptation methods. Thus we use two popular edge development platforms Raspberry Pi (Model 4B) and NVIDIA Jetson Orin to evaluate the practicality of deploying on-device TTA on a CPU and a GPU-powered edge device. We monitor the following:

\begin{itemize}
\item \textbf{CPU\slash GPU Consumption}: The percentage of CPU\slash GPU utilization during model adaptation across different TTA methods.
\item \textbf{Memory Consumption}: The amount of memory used by the adaptation process, highlighting memory efficiency across methods.

\end{itemize}

\subsection{Datasets and Setup}
Here we detail our selection of datasets, models, and their evaluation strategies and argue their choices.

\subsubsection{Dataset and distribution shifts}

\cifarc~\cite{hendrycks2019robustness} is a well-known synthetic dataset that introduces controlled corruption to the original \cifar~\cite{krizhevsky2009learning} image recognition dataset, enabling the study of model robustness under distribution shifts. It contains samples from the same $10$ categories as the \cifar{} dataset. It consists of $15$ types of distribution shifts, each representing a corruption that simulates a real-world scenario. Each type of shift includes $5$ severity levels of corruption with level $5$ denoting the most corrupted.  Although these corruptions are synthetically generated (using certain random distributions like Gaussian), they are designed to closely resemble distortions that an edge device might encounter, such as noise, blur, or changes in brightness. For each corruption type, \cifarc{} contains $10,000$ samples with $1,000$ samples from each of the $10$ categories.

\pacs~\cite{yu2022pacs} is a widely used image classification dataset used in the domain adaptation community. It consists of four domains: Photo ($1,670$ images), Art Painting ($2,048$ images), Cartoon ($2,344$ images) and Sketch ($3,929$ images). Unlike \cifarc{}, \pacs{} comprises real-world images collected from different sources with vastly different visual characteristics. The dataset consists of seven categories, making it a relatively small yet diverse benchmark for evaluating model robustness across domain shifts. \pacs{} is imbalanced in multiple ways. The number of samples per domain varies significantly, and within each domain, the number of samples per category is also imbalanced.

In line with popular choices in TTA algorithmic studies, we limit ourselves to the above two image classification tasks and leave evaluating other modalities for the future.

\subsubsection{Selecting the TTA methods to benchmark}
From the brief survey of recent TTA (See Section~\ref{sec:background}) approaches, we categorize them into four categories as summarised in \tablename~\ref{tbl:botta_tta}. We take one representative work from each category, which will be thoroughly assessed in BoTTA considering data and edge-device constraints.
\begin{table}[]
\centering
\caption{TTA algorithms used in BoTTA. We choose one of the SOTA algorithms from each category.}
\label{tbl:botta_tta}
\resizebox{\columnwidth}{!}{%
\begin{tabular}{|l|l|}
\hline
\textbf{Algorithmic Approach}        & \textbf{Key Works} \\ \hline
Softmax Entropy Minimization         & TENT~\cite{wang2020tent}               \\ \hline
Sharpness Aware Entropy Minimization & SAR~\cite{niu2023towards}          \\ \hline
Pseudo-Labeling based                & SHOT~\cite{liang2020we}              \\ \hline
Instance-Aware Batch Normalization                & NOTE~\cite{gong2023note}              \\ \hline
Optimization-free                    & T3A~\cite{iwasawa2021test}                \\ \hline
\end{tabular}%
}
\end{table}
\subsubsection{Model architectures.}

In this work, we leverage pre-trained models for our experiments, specifically \resnet{}, \resnetF{}, and \vit{} (vit-small-patch$16$-$224$). These architectures have been widely adopted in the literature for benchmarking robustness and adaptation. We use the pre-trained model parameters provided by~\cite{zhao2023pitfalls}. Readers may refer to the original paper for a detailed exposition of pre-training methodologies. While both \resnet{} and \resnetF{} follow the standard deep residual learning framework, they differ in the depth of layers. Vision Transformer (ViT) uses inductive biases with self-attention mechanisms offering improved performance on large-scale datasets. It requires significantly more computational power, making it impractical for edge devices. By leveraging these pre-trained architectures, we ensure consistency with previous benchmarking efforts while facilitating robust evaluations under various TTA settings.

\subsubsection{Model evaluation strategy}\label{sec:model-eval-strategy}
Typically TTA algorithms are evaluated in the following way. The models are adapted and evaluated at the same time. Most of the TTA methods need to compute gradients through perform memory-and compute-intensive back-propagation, hindering real-time inference and being costly for resource constrained devices.

Instead, the models at the edge devices can be adapted periodically. This strategy will save critical resources and enable more efficient inference. The device will use the frozen (adapted) model ($\theta_{\Delta_T}$) to perform inference on the forthcoming samples. Additionally, the device will buffer the target data samples to be used in the next opportunity for adaptation. The periodicity of the adaptation will be driven by two factors: firstly, the availability of resources for model adaptation, and secondly, the size of the sample buffer.

We further refrain from updating the batch normalization parameters based on the current batch of samples at test time which is a common practice in TTA methods such as TENT~\cite{wang2020tent}. This is because such statistics quickly overfit the current batch especially when the batch size is small. While using small batch sizes is a popular way to reduce memory requirements in edge devices, we freeze all model parameters during evaluations. We further save the adapted model in the device file storage and then load it in the primary memory, following the usual way to handle unexpected power failures in the edge devices. Due to these settings, our results show lower accuracy numbers compared to other TTA literature.
\begin{figure}[]
    \centering
    \begin{minipage}{0.48\columnwidth}
        \centering
        \includegraphics[width=\columnwidth,keepaspectratio]{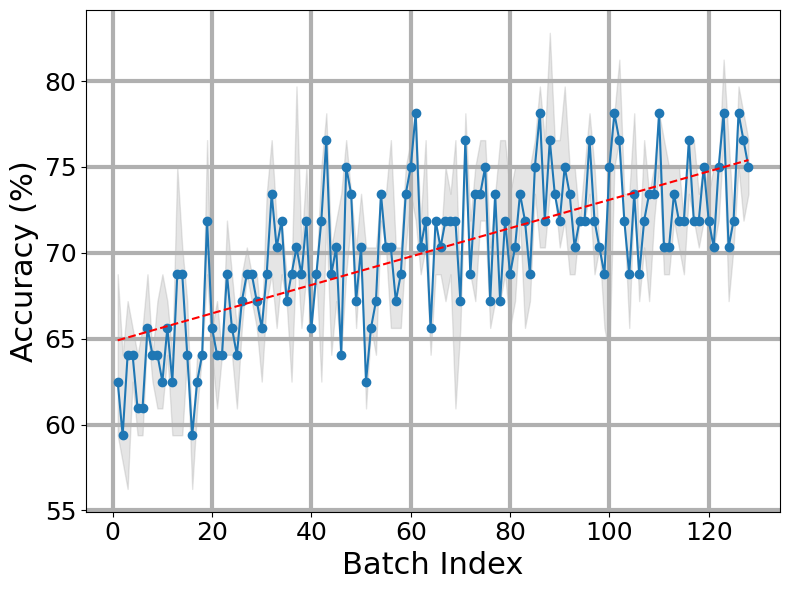}
        \subcaption{}
        \label{fig:batch-acc-a}
    \end{minipage}%
    \hfill
    \begin{minipage}{0.48\columnwidth}
        \centering
        \includegraphics[width=\columnwidth,keepaspectratio]{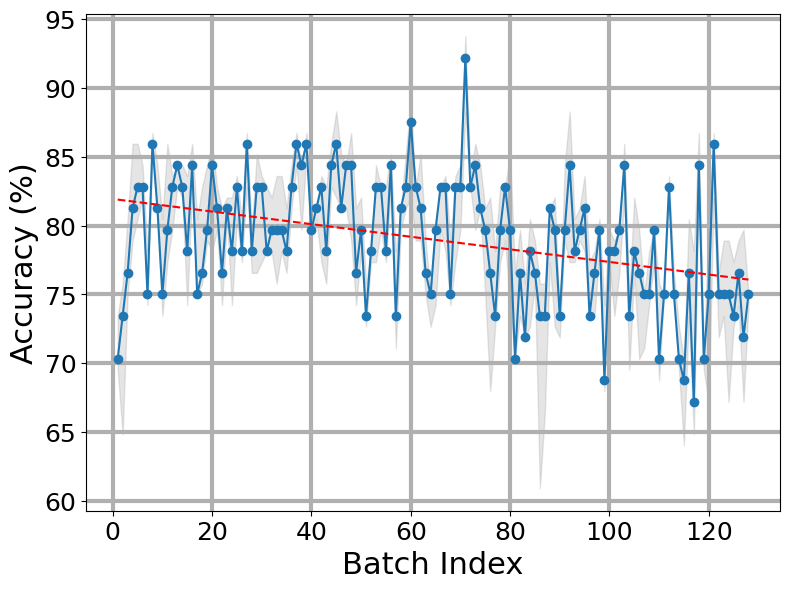}
        \subcaption{}
        \label{fig:batch-acc-b}
    \end{minipage}
    \caption{Test accuracy on the batches during adaptation of \cifarc{} Gaussian noise domain (severity $5$) with SHOT method using (a) \resnet{} and (b) \vit.}
    \label{fig:batch-acc}
\end{figure}

As a frozen model will be used for inference, it is important to evaluate its generalization. We thus perform this in three test scenarios: $i)$ $\Delta_T$, $ii)$ $D_T$, and $iii)$ an out of distribution to $\Delta_T$, i.e., $\Delta_{OOD}$. While $(i)$  aligns with the currently used evaluation strategies in the literature, the rest of the two scenarios are especially relevant to edge device settings due to inference using frozen adapted models. In line with TTA literature, we keep the batch size at $64$ in all experiments.

To ensure robustness, each experiment in this paper is executed $5$ times (except Section~\ref{sec:on_device}) with independent seed selection, and every result presented here is the median.

\myparagraph{Selecting the model to evaluate.} As continuous adaptation is infeasible, a critical question thus is which state of the model should be stored for future inferences.

In this paper, we choose to store the model produced at the end of the adaptation, i.e., after it processes all the data samples in $\Delta_T$. As batch size used during adaptation severely affects the adapted model performance~\cite{zhao2023pitfalls}, wherever possible, we avoid such catastrophic performance drop by choosing a suitable $|\Delta_T|$ such that the last batch is always full.

This is in line with the observations in \figurename~\ref{fig:batch-acc}(a) where the test accuracy goes up as the model processes more batches of data. While the test accuracy varies because the evaluation set (the batch) is small ($64$), the upward trend (fitted by a linear regression) is clear. However, in the case of ViT in \figurename~\ref{fig:batch-acc}(b), we see the model accuracy falls when it processes more batches, however, we keep the same selection strategy irrespective of the model architecture used. While smarter model selection strategies exist~\cite{zhao2023pitfalls}, they also cost memory (e.g., to store different versions of the adapted model) and compute thus we leave their investigation for the future.
\begin{figure*}[]
    \centering
    \includegraphics[keepaspectratio=true, width=0.7\linewidth]{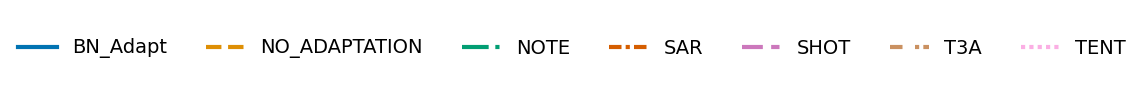}
    \begin{minipage}{0.24\textwidth}
        \centering
        \includegraphics[width=\columnwidth,keepaspectratio]{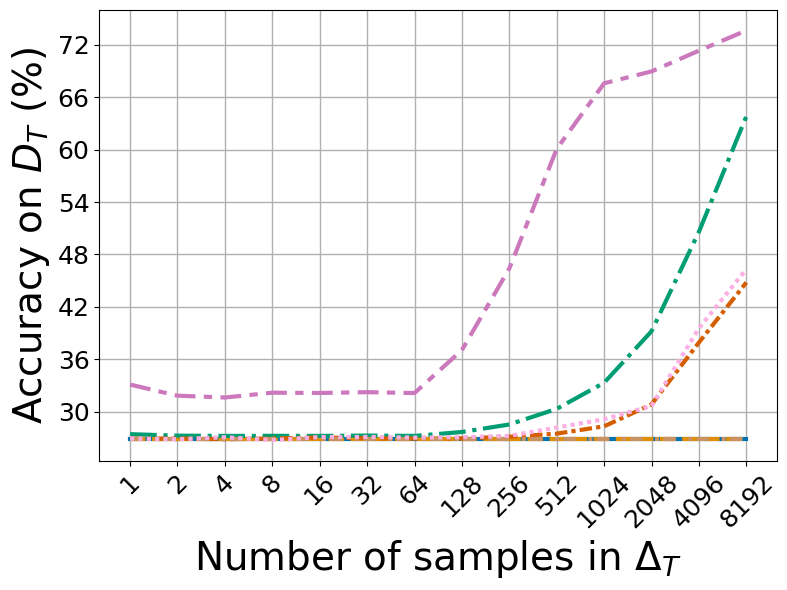}
        \subcaption{}
        \label{fig:S1-resnet-fog-gaussian-a}
    \end{minipage}%
    \begin{minipage}{0.24\textwidth}
        \centering
        \includegraphics[width=\columnwidth,keepaspectratio]{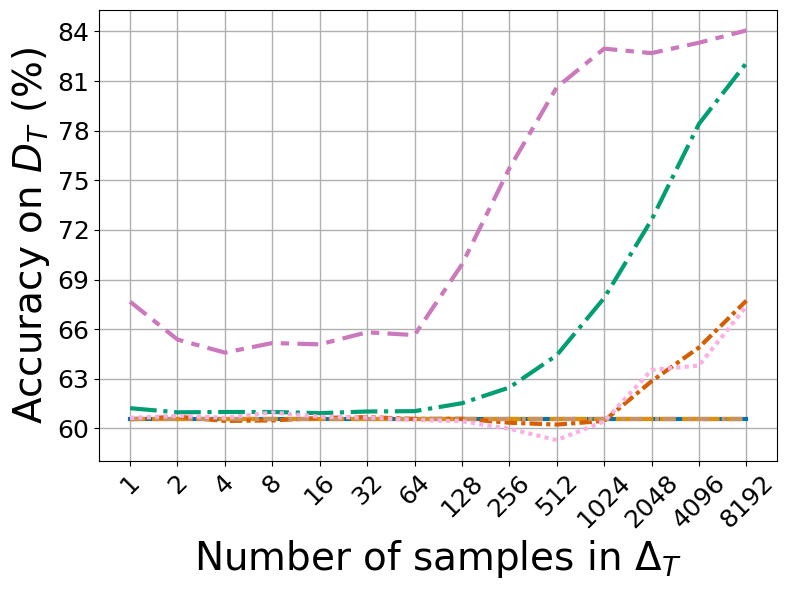}
        \subcaption{}
        \label{fig:S1-resnet-fog-gaussian-b}
    \end{minipage}%
    \begin{minipage}{0.24\textwidth}
        \centering
        \includegraphics[width=\columnwidth,keepaspectratio]{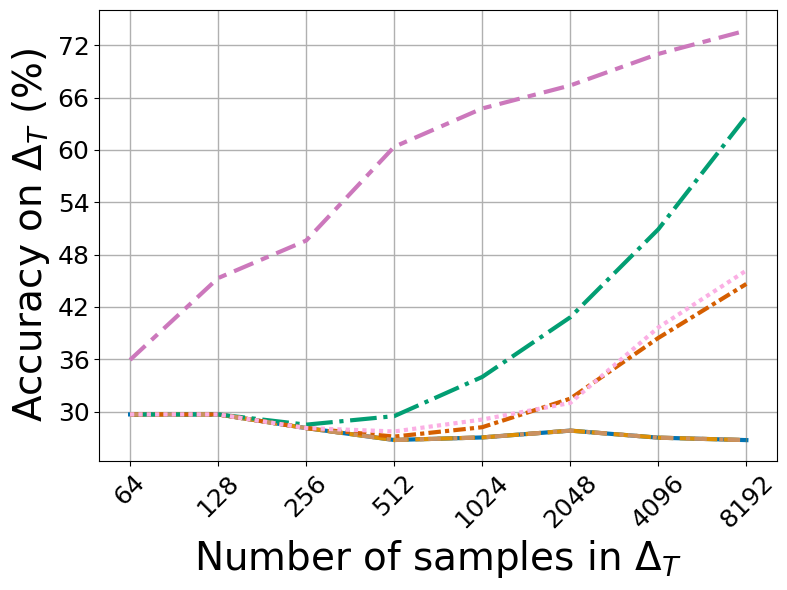}
        \subcaption{}
        \label{fig:S1-resnet-fog-gaussian-c}
    \end{minipage}%
    \begin{minipage}{0.24\textwidth}
        \centering
        \includegraphics[width=\columnwidth,keepaspectratio]{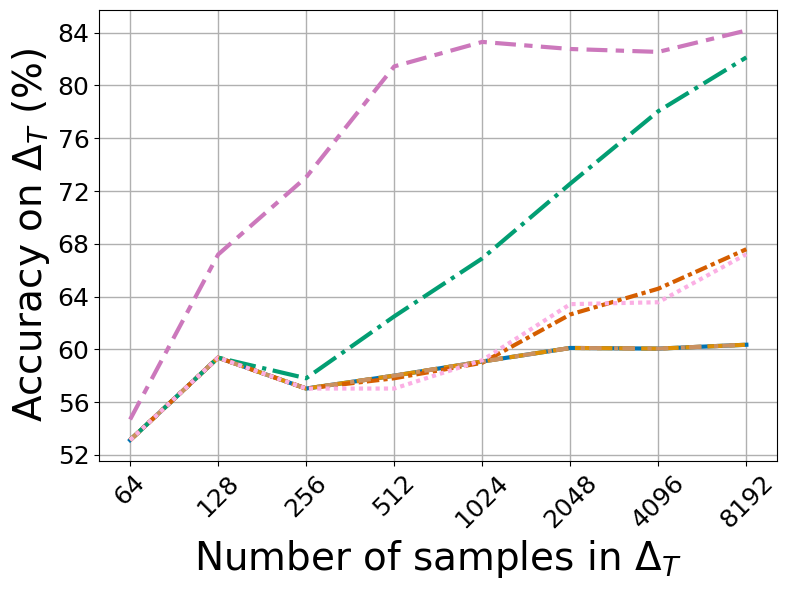}
        \subcaption{}
        \label{fig:S1-resnet-fog-gaussian-d}
    \end{minipage}
    \caption{Test accuracy on $D_T$ with \cifarc{} Gaussian Noise (a) and Fog (b) using \resnet{} architecture. (c) and (d) show accuracy on Gaussian noise and Fog, respectively, when testing using $\Delta_T$. Both corruption types have a severity of $5$. While accuracy increases with the increasing $|\Delta_T|$, with data size below $64$, none of the TTA methods works well. SHOT consistently remains the most accurate across all settings here.}
    \label{fig:S1-resnet-fog-gaussian}
\end{figure*}
\section{Observations on Scenario $\mathbf{1}$: Limited Target Samples}
\label{sec:scenario1}
Here, we evaluate the TTA algorithms on the scenario $1$ defined in Section~\ref{sec:scenario-def-1}, where various target samples are available for adaptation. We present an exhaustive suite of settings covering all model architectures (\resnet{}, \resnetF{} and \vit) and both datasets (\cifarc{} and \pacs).

\myparagraph{Observations on \cifarc.} We report the accuracy numbers using \resnet{} followed by \vit. This scenario considers the adaptation and its evaluation using the same domain. Here, the observations are made using two domains: Gaussian noise and Fog, where both use their highest severity level ($5$). Though both are synthetically generated in \cifarc{}, these two domains cover a representative set with the most popular noise used in the literature and a realistic corruption, respectively.

\myparagraph{Results on \resnet.}
The following observations are on a \resnet{} model that is pre-trained on the \cifar{} dataset and is adapted on $\Delta_T$ consisting of randomly selected samples from $D_T$. SHOT consistently outperforms across all values of $|\Delta_T|$ tested in \figurename~\ref{fig:S1-resnet-fog-gaussian} (a-d). The following tests on two test datasets, namely, \figurename~\ref{fig:S1-resnet-fog-gaussian} (a, b) use of all $10k$ samples from the same domain as $\Delta_T$ and \figurename~\ref{fig:S1-resnet-fog-gaussian} (c, d) use $\Delta_T$.

\myparagraph{Evaluating on $D_T$.} Across all the settings in \figurename~\ref{fig:S1-resnet-fog-gaussian}, we observe that TENT and SAR behave similarly. Even though the accuracy remains low with a large dataset with $|\Delta_T| = 8192$ ($46.18\%$ on the Gaussian noise domain), notably, they show no model improvement with $|\Delta_T| \le 512$ in both the corruption types.

The backpropagation-free adaptation method, T3A, cannot manifest accuracy gains over the source model. This shows that while it can improve accuracy when the data distribution shift is mild, their simple classifier adjustment policy is insufficient when the shift is severe. 

While with small adaptation data, $|\Delta_T| \le 64$, the accuracy of SHOT does not improve, a steady linear improvement is shown with $|\Delta_T| \ge 128$ (\figurename~\ref{fig:S1-resnet-fog-gaussian} (a, b)). Notably, in both corruption types, the model accuracy improves over $\theta_S$ even when $|\Delta_T| = 1$; this is attributed to the optimization using pseudo-labeling.

\myparagraph{Evaluating on $\Delta_T$.} While the essential patterns remain the same between \figurename~\ref{fig:S1-resnet-fog-gaussian} (c, d) and (a, b), here with smaller $\Delta_T$, the absolute accuracy values are naturally higher compared to \figurename~\ref{fig:S1-resnet-fog-gaussian} (a, b). For example, SHOT achieves $37.05\%$ and $45.31\%$ accuracy with $|\Delta_T| = 128$ on Gaussian noise when using $D_T$ and $\Delta_T$ respectively. This is because the adaptation and test data are the same, which shows that the typical way of evaluating the TTA methods (on $\Delta_T$) overestimates the model performance for small $\Delta_T$. We omit the portion of the curves below $|\Delta_T| = 64$ in \figurename~\ref{fig:S1-resnet-fog-gaussian} (c, d) because of the high uncertainty with a small test data.

\begin{figure}[]
    \centering
    \includegraphics[keepaspectratio=true, width=0.5\linewidth]{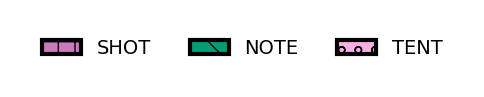}
    
    \begin{minipage}{0.48\columnwidth}
        \centering
        \includegraphics[width=\columnwidth,keepaspectratio]{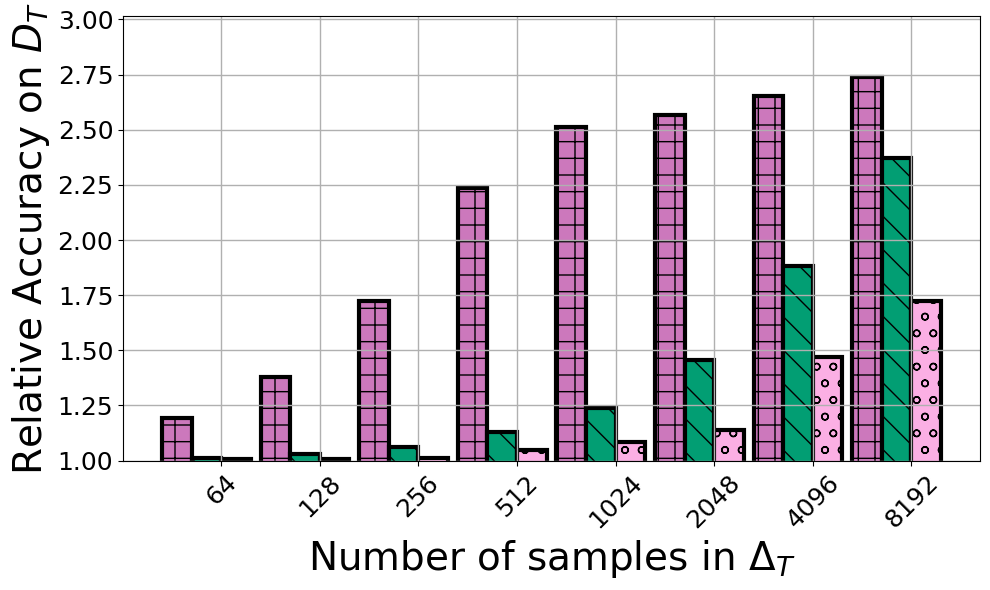}
        \subcaption{}
        \label{fig:S1-rag-resnet-fog-gaussian-a}
    \end{minipage}%
    \begin{minipage}{0.48\columnwidth}
        \centering
        \includegraphics[width=\columnwidth,keepaspectratio]{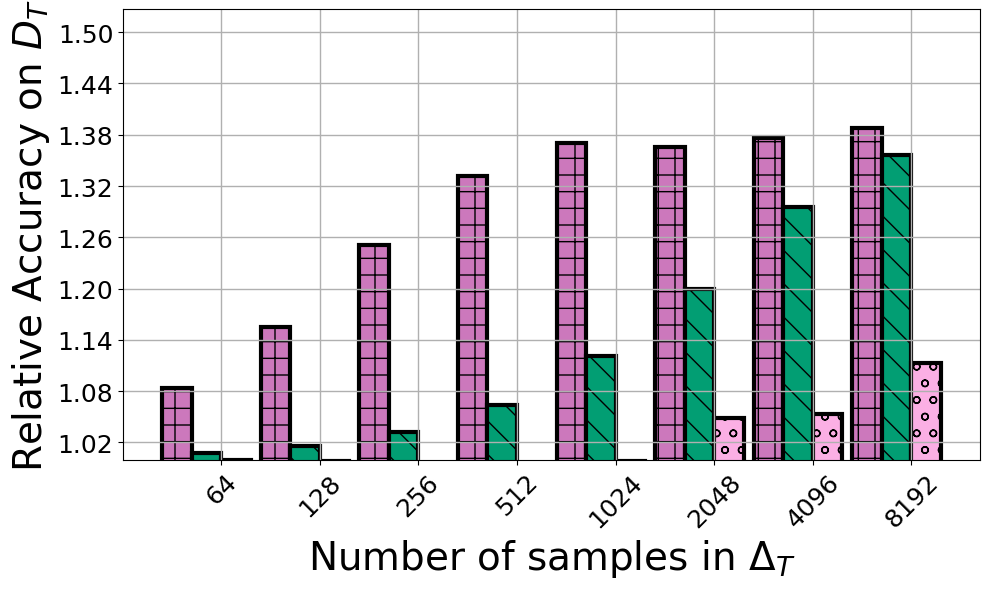}
        \subcaption{}
        \label{fig:S1-rag-resnet-fog-gaussian-b}
    \end{minipage}
    \caption{Relative accuracy gain (\resnet) using different TTA methods compared to $\theta_S$ while the target domains are (a) Gaussian Noise (b) Fog in \cifarc{}.}
    \label{fig:S1-rag-resnet-fog-gaussian}
\end{figure}

\myparagraph{Relative accuracy gains.} The accuracy gains for different TTA methods (\figurename~\ref{fig:S1-rag-resnet-fog-gaussian}) are computed as a fraction compared to the source model's ($\theta_S$) accuracy, i.e., $\frac{\xi(\theta_{\Delta_T}, D_T)}{\xi(\theta_S, D_T)}$. Given that the absolute model accuracy is affected heterogeneously in various domains, this provides a way to compare accuracy gains across different domains. As $\theta_S$ has massively different source accuracy across Fog and Gaussian noise ($60.55\%$ and $26.89\%$ respectively), the accuracy gains with the Fog domain remain lower than the Gaussian noise, namely SHOT achieves $2.74\times$ and $1.39\times$ gain respectively when using $|\Delta_T| = 8192$. However, the trend of increasing the accuracy gain with $|\Delta_T|$ until reaching a plateau remains similar across the two domains.

\myparagraph{Results on \vit.} Here, a Vision Transformer (\vit) model is pre-trained on the \cifar{} dataset and adapted to Gaussian noise domain in \cifarc{}.

\begin{figure}[]
    \centering
    \includegraphics[keepaspectratio=true, width=0.9\linewidth]{final_figures/legend.png}
    
    \begin{minipage}{0.48\columnwidth}
        \centering
        \includegraphics[width=\columnwidth,keepaspectratio]{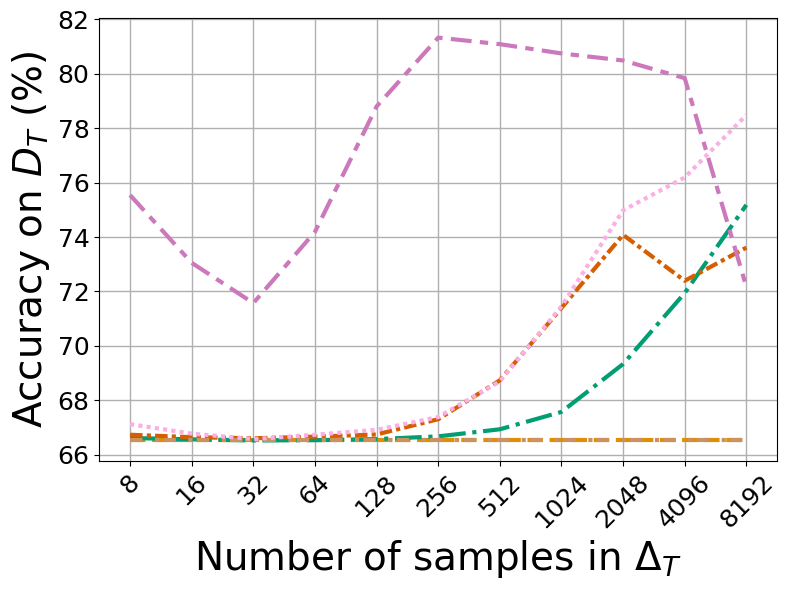}
        \subcaption{(a)}
        \label{fig:S1-ViT-a}
    \end{minipage}%
    \hfill
    \begin{minipage}{0.48\columnwidth}
        \centering
        \includegraphics[width=\columnwidth,keepaspectratio]{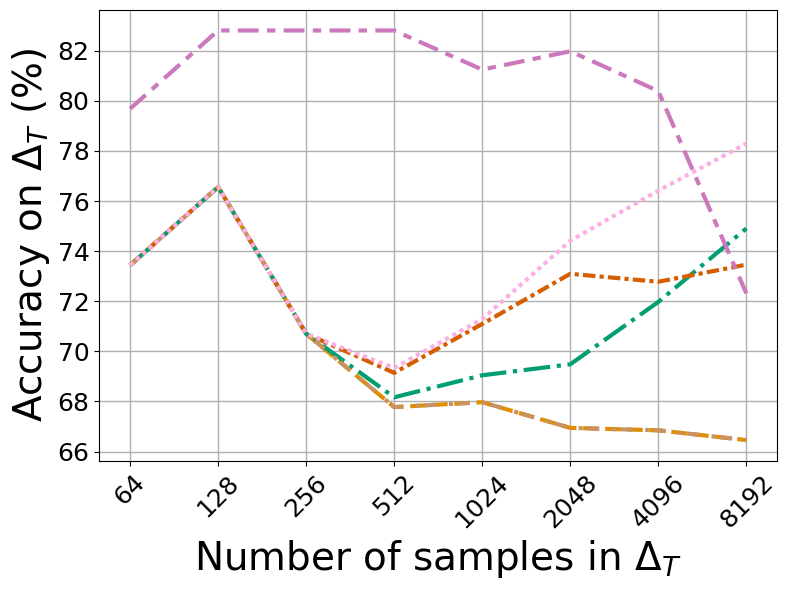}
        \subcaption{(b)}
        \label{fig:S1-ViT-b}
    \end{minipage}

    \caption{Shows test accuracy on $D_T$ and $\Delta_T$ in (a) and (b) respectively using \vit{} while adapting on Gaussian noise in \cifarc{}. Results follow a trend similar to \figurename~\ref{fig:S1-resnet-fog-gaussian}: none of the methods show a significant improvement in the low data regime. Surprisingly the accuracy for SHOT declines with $|\Delta_T| > 4096$.}
    \label{fig:S1-ViT}
\end{figure}

\begin{figure*}[]
    \centering
    \begin{minipage}{0.355\textwidth}
        \centering
        \includegraphics[width=\columnwidth,keepaspectratio]{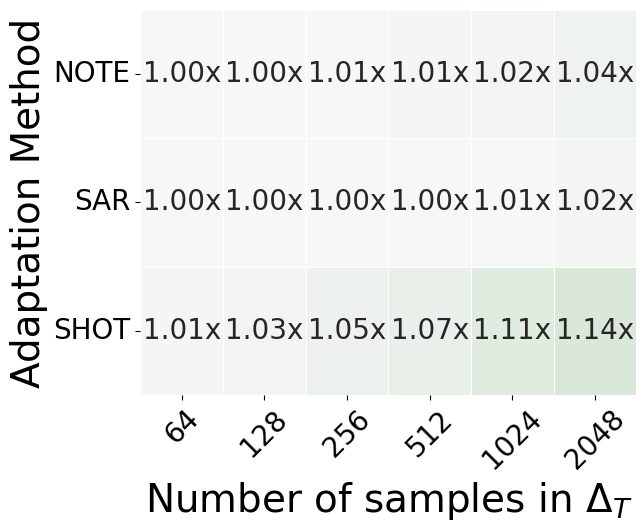}
        \subcaption{}
        \label{fig:S1-pacs-rag-a}
    \end{minipage}%
    \begin{minipage}{0.29\textwidth}
        \centering
        \includegraphics[width=\columnwidth,keepaspectratio]{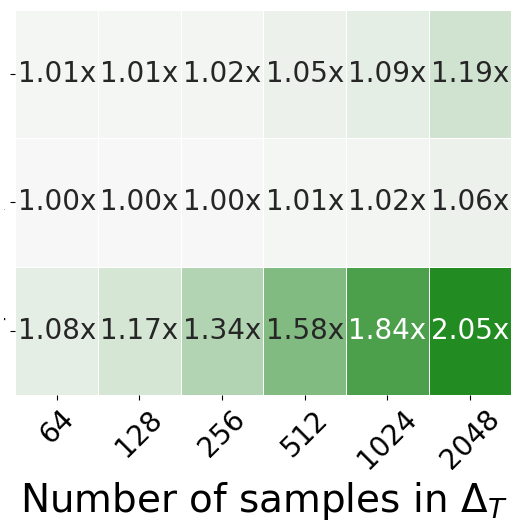}
        \subcaption{}
        \label{fig:S1-pacs-rag-b}
    \end{minipage}%
    \begin{minipage}{0.355\textwidth}
        \centering
        \includegraphics[width=\columnwidth,keepaspectratio]{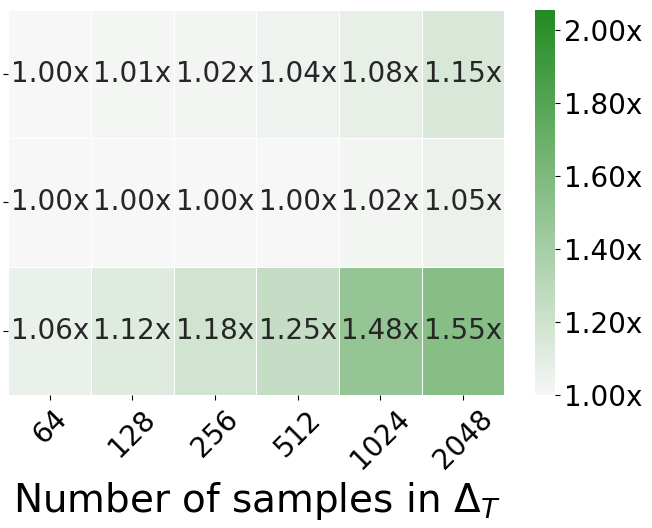}
        \subcaption{}
        \label{fig:S1-pacs-rag-c}
    \end{minipage}
    \caption{Relative accuracy gain using different TTA methods compared to the source model's ($\theta_S$) accuracy across a varying number of samples in adaptation data ($\Delta_T$) on \pacs{} domains (a) Art, (b) Cartoon, and (c) Sketch using \resnetF{} architecture pre-trained on the domain: `Photo'.}
    \label{fig:S1-pacs-rag}
\end{figure*}

Results in \figurename~\ref{fig:S1-ViT} reinforce that model architecture significantly affects the adaptation accuracy. For example, when using $|\Delta_T| = 8192$, TENT achieves $78.48\%$ (\figurename~\ref{fig:S1-ViT} (a)) and $46.33\%$ (\figurename~\ref{fig:S1-resnet-fog-gaussian} (a)) while using \vit{} and \resnet{} respectively. However, other trends remain similar to the \resnet{} architecture; for example, SHOT is the most accurate TTA method (when $|\Delta_T| \le 4096$ and none of the methods improve the model with a small $\Delta_T$.

TENT improves the models with $|\Delta_T| > 4096$ and achieves the best result, improving the source model accuracy by $1.15\times$. NOTE follows a similar trend as in \resnet{} experiments. T3A fails to improve the model accuracy, denoting that only adjusting the classifier is insufficient even with a more sophisticated encoder like \vit{}.

Accuracy for SHOT decreases with small values of $|\Delta_T| \le 32$, and then it achieves a linear increase. Further, its accuracy drops significantly from $80\%$ to $72\%$ with $|\Delta_T| = 4096$ and $8192$ respectively. While this requires future investigation, we believe this aligns with the fact that the accuracy of SHOT decreases in the \vit{} model when it processes more batches, as shown in \figurename~\ref{fig:batch-acc} (b).

\myparagraph{Observations on \pacs.}
\figurename~\ref{fig:S1-pacs-rag} compares the accuracy gains relative to the source model $\theta_S$ of three TTA methods, including NOTE, SHOT and SAR, that have the highest accuracy gains with \cifarc{}. Observations here are consistent with the \cifarc{} results above: SHOT produces the highest accuracy across all the settings in \figurename~\ref{fig:S1-pacs-rag} (a-c) and SAR performs the worst (improving at most $1.06\times$). Compared to that, SHOT achieves an accuracy gain of up to $2.05\times$. Again, similar to the results with \cifarc{}, we observe that none of the methods achieve accuracy gain in low $|\Delta_T|$ regime; for example, with $|\Delta_T|=128$ samples, the accuracy gain is up to $1.17\times$ across the scenarios.

\section{Observations on Scenario 2: Limited Categories to Adapt}
\label{sec:scenario2}

Here, we evaluate the TTA algorithms on Scenario $2$ as defined in Section~\ref{sec:S2-defn}. We randomly select $k$ categories to be used for adaptation. Then, we randomly choose $960$ samples from each selected category when using \cifarc{} and $124$ samples when using the \pacs{} dataset to be included in $\Delta_T$. Thus, as a side effect, the size of $\Delta_T$ also increases with the number of categories.

Each experiment is run for $5$ times the categories, and the samples are chosen apriori for these $5$ experiments, i.e., the $\Delta_T$ is fixed for each run. The only stochastic element between different runs is the batch selection between different executions. As a side effect, there is a surprising but insightful manifestation of such data sampling across experiments, as detailed in the appropriate places below. In line with the definitions in Section~\ref{sec:S2-defn}, the adapted model is evaluated in three test settings as below.
\begin{figure*}[]
    \centering
    \includegraphics[keepaspectratio=true, width=0.7\linewidth]{final_figures/legend.png}
    
    \begin{minipage}{0.24\textwidth}
        \centering
        \includegraphics[width=\linewidth,keepaspectratio]{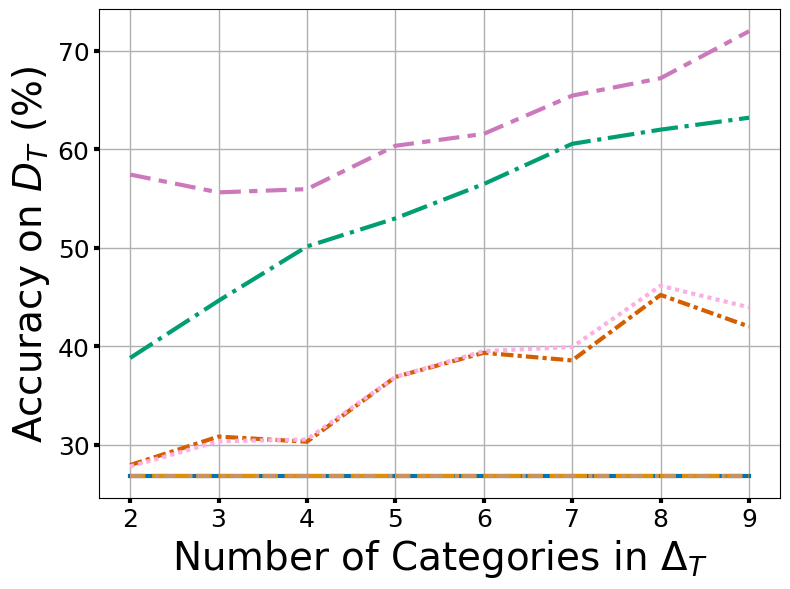}
        \subcaption{(a)}
        \label{fig:S2-a}
    \end{minipage}%
    \begin{minipage}{0.24\textwidth}
        \centering
        \includegraphics[width=\linewidth,keepaspectratio]{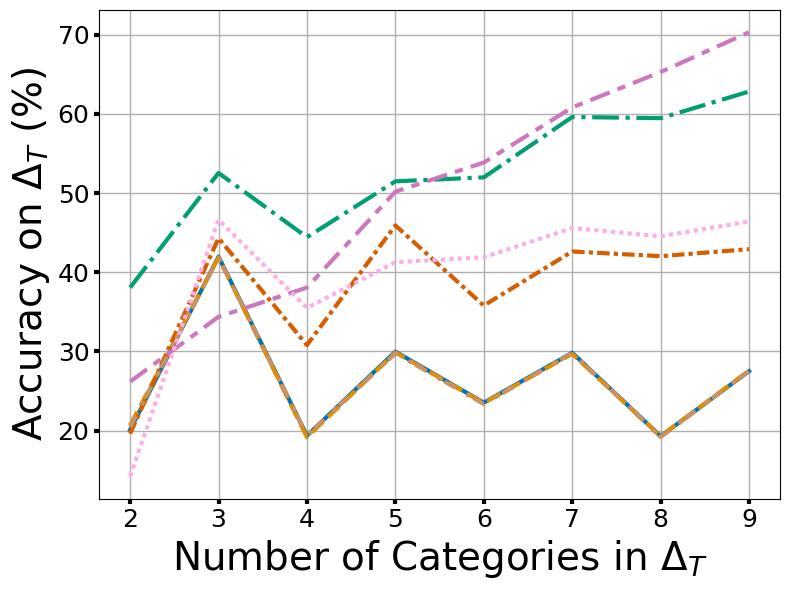}
        \subcaption{(b)}
        \label{fig:S2-b}
    \end{minipage}%
    \begin{minipage}{0.24\textwidth}
        \centering
        \includegraphics[width=\linewidth,keepaspectratio]{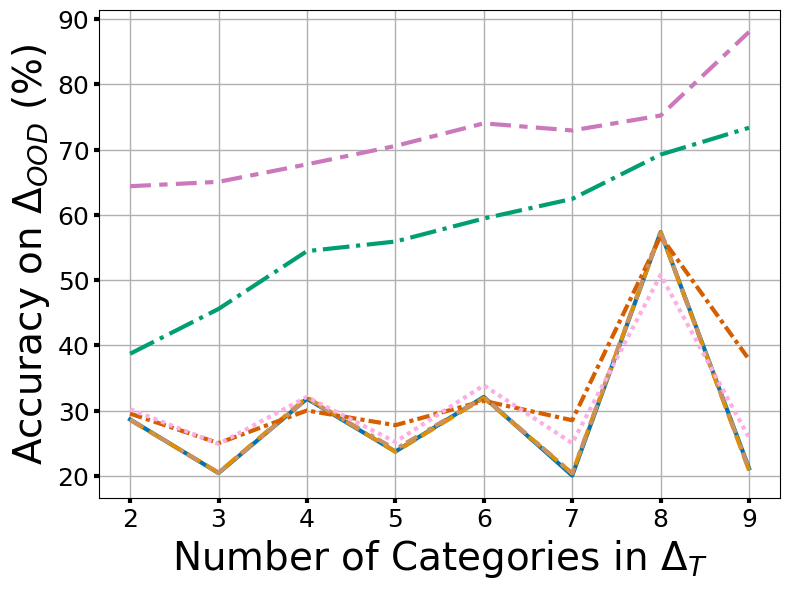}
        \subcaption{(c)}
        \label{fig:S2-c}
    \end{minipage}%
    \begin{minipage}{0.24\textwidth}
        \centering
        \includegraphics[width=\linewidth,keepaspectratio]{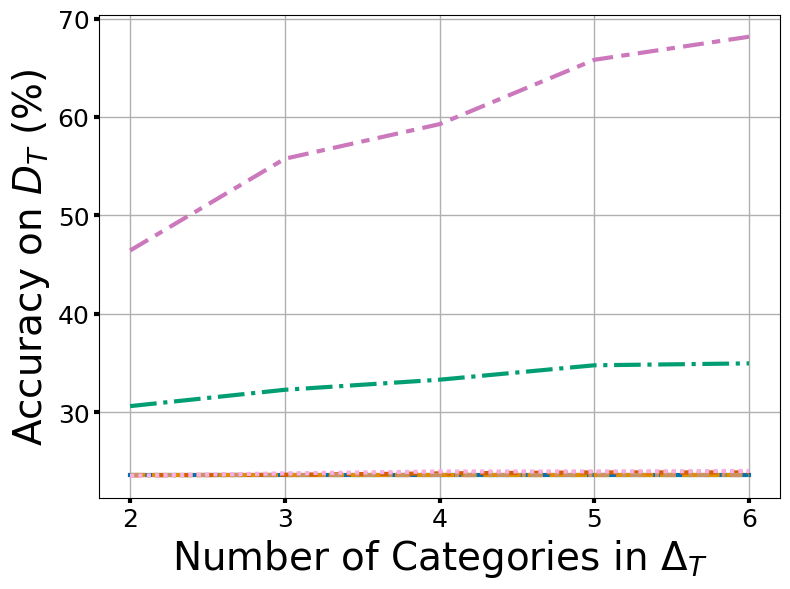}
        \subcaption{(d)}
        \label{fig:S2-d}
    \end{minipage}
    \caption{Comparing TTA methods when using a subset of categories for adaptation. (a), (b), and (c) show results from the \cifarc{} dataset while using Gaussian noise corruption with severity $5$. In contrast, (d) uses \pacs{} dataset to transfer knowledge from the Sketch domain to the Art domain. The evaluation sets align with Section~\ref{sec:S2-defn}. Test accuracy improves across all settings as $|\mathcal{C}(\Delta_T)|$ increases. SHOT remains most accurate across all settings (a-d) except in (b) where NOTE has better accuracy when $|\mathcal{C}(\Delta_T)| \le 5$. There are unusual accuracy fluctuations in (b) and (c) explained by the choice of the classes in $\Delta_T$.}
    \label{fig:S2}
\end{figure*}

\myparagraph{Evaluating on $D_T$.} We show results for both CIFAR-10C (\figurename~\ref{fig:S2}(a)) and \pacs{} (\figurename~\ref{fig:S2}(d)) for this setting. In the case of \cifarc{}, we use Gaussian noise corruption with a severity of $5$. Here, \figurename~\ref{fig:S2}(a) evaluates an adapted model based on $10k$ samples in \cifarc{} Gaussian noise domain (sev $5$) which contains all $10$ classes in the dataset. \figurename~\ref{fig:S2}(a) shows that the accuracy of the different methods improves as $|\mathcal{C}(\Delta_T)|$ increases. 

SHOT consistently performs the best across the two datasets. It reaches over $70\%$ for \cifarc{} with $|\mathcal{C}(\Delta_T)|=9$ and $68\%$ for \pacs{} dataset with $|\mathcal{C}(\Delta_T)|=6$. Remarkably, in \cifarc{} SHOT produces $57.44\%$ accuracy with only $|\mathcal{C}(\Delta_T)|=2$ improving from the source model accuracy of $26.89\%$. The results in \pacs{} follow a similar trend where the SHOT method achieves an accuracy of $46.43\%$ when $|\mathcal{C}(\Delta_T)|=2$. With increasing $|\mathcal{C}(\Delta_T)|$, the accuracy typically increases in all the methods, denoting that the methods need representations from all categories to generalize in those categories.
\begin{figure}[]
    \centering
    \includegraphics[width=.7\columnwidth]{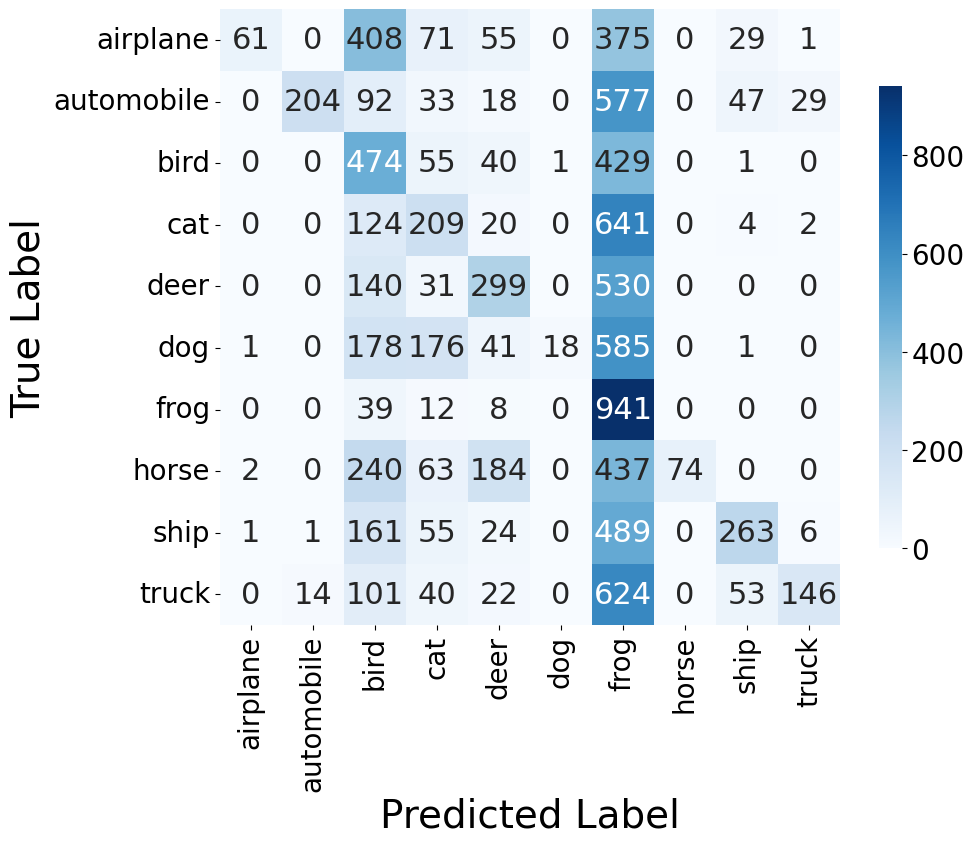}
    \caption{Confusion matrix for \resnet{} trained on \cifar{} and evaluated on the \cifarc{} Gaussian noise domain (severity $5$). Surprisingly many samples are classified as `Frog'.
    }
    \label{fig:conf_matrix}
\end{figure}
TENT and SAR perform significantly differently across \cifarc{} and \pacs{}. While they fail to improve the accuracy in \pacs{} (\figurename~\ref{fig:S2} (d)), they improve accuracy to $43.96\%$ and $42.01\%$ respectively in \cifarc{} (\figurename~\ref{fig:S2} (a)) compared to $26.89\%$ accuracy in $\theta_S$. The improvement is attributed to our sampling method, where an increase in  $|\mathcal{C}(\Delta_T)|$ also increases the number of samples in $\Delta_T$. As the increase happens at a higher rate in \cifarc{} compared to \pacs{} due to the number of samples chosen from a category, the accuracy improvement is more pronounced in \cifarc{}.

\myparagraph{Evaluating on $\Delta_T$.} \figurename~\ref{fig:S2}(b) follows a similar trend as \figurename~\ref{fig:S2}(a), that the accuracy improves with increasing $|\mathcal{C}(\Delta_T)|$. NOTE produces the most accurate model with small $|\mathcal{C}(\Delta_T)| \le 5$ whereas SHOT becomes the best method with $|\mathcal{C}(\Delta_T)| > 5$.

There is a surprising spike in accuracy at $|\mathcal{C}(\Delta_T)|=3$ in \figurename~\ref{fig:S2}(b). The reason behind this is the selection of the exact categories in $\mathcal{C}(\Delta_T)$. As a pre-trained \resnet{} model predicts a disproportionate number of samples to be a `Frog' (\figurename~\ref{fig:conf_matrix}) if the 'Frog' category is included in $\mathcal{C}(\Delta_T)$ then the model accuracy falsely improves. Though we randomly sampled $3$ classes out of $10$ categories, the particular random choice selected the `Frog' class in $4$ out of the $5$ runs, resulting in a higher accuracy on $\Delta_T$.

\begin{figure*}[]
    \centering
    \includegraphics[keepaspectratio=true, width=0.7\linewidth]{final_figures/legend.png}
    
    \begin{minipage}{0.24\textwidth}
        \centering
        \includegraphics[width=\columnwidth,keepaspectratio]{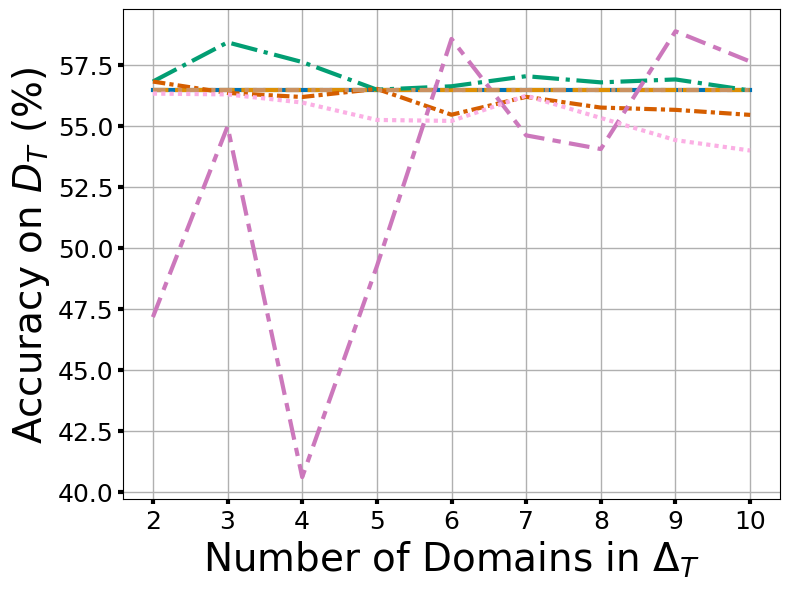}
        \subcaption{}
        \label{fig:S3-a}
    \end{minipage}%
    \begin{minipage}{0.24\textwidth}
        \centering
        \includegraphics[width=\columnwidth,keepaspectratio]{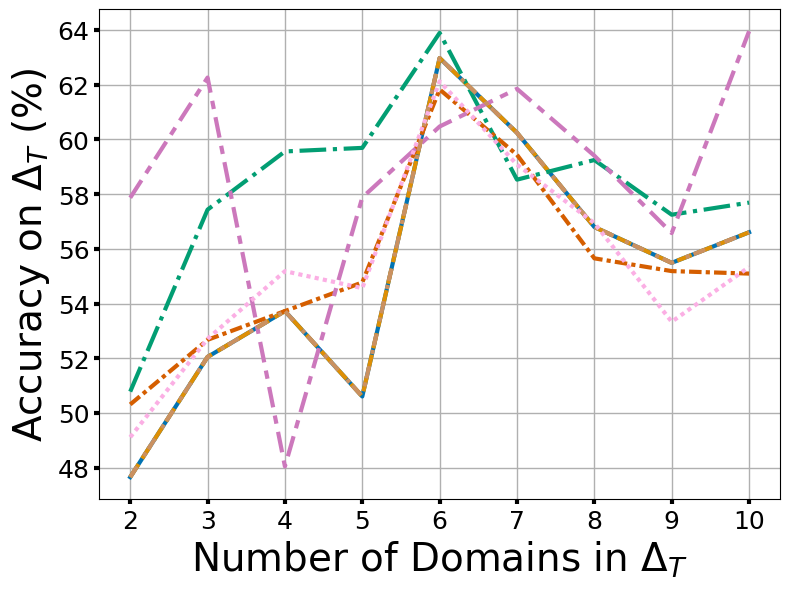}
        \subcaption{}
        \label{fig:S3-b}
    \end{minipage}%
    \begin{minipage}{0.24\textwidth}
        \centering
        \includegraphics[width=\columnwidth,keepaspectratio]{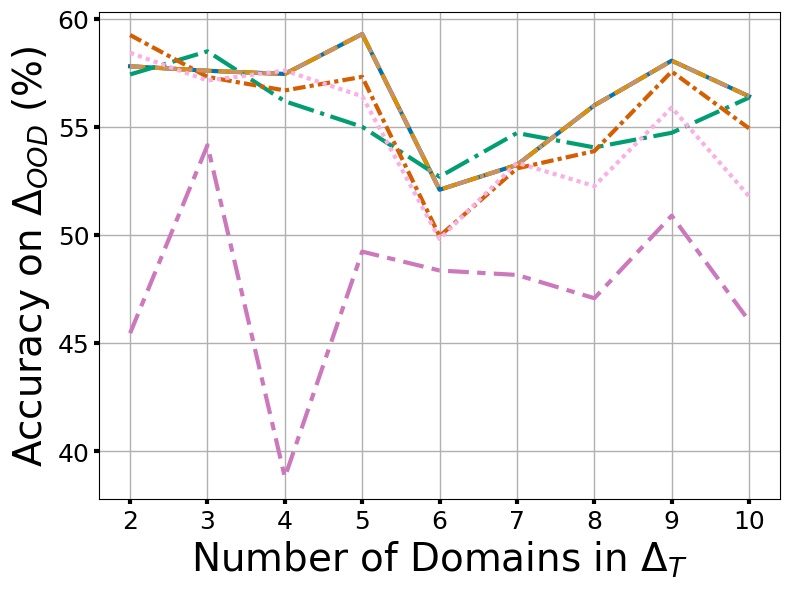}
        \subcaption{}
        \label{fig:S3-c}
    \end{minipage}%
    \begin{minipage}{0.24\textwidth}
        \centering
        \includegraphics[width=\columnwidth,keepaspectratio]{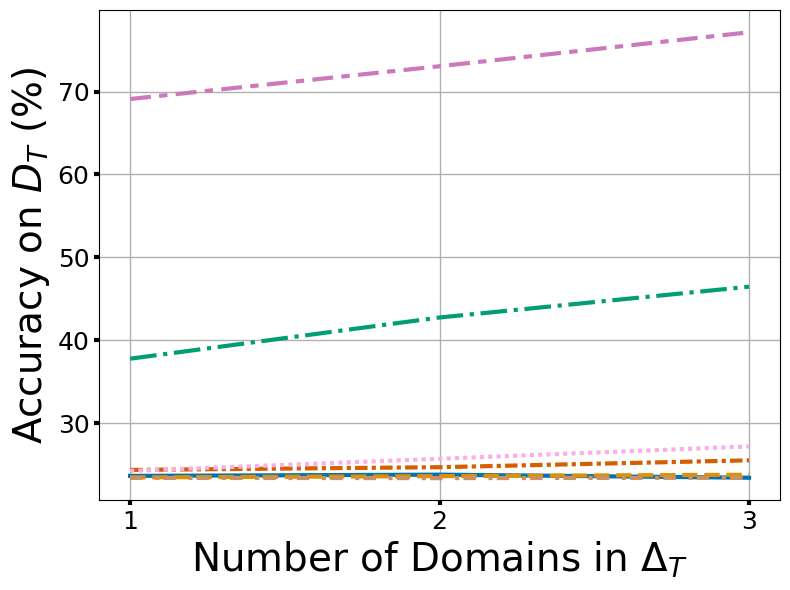}
        \subcaption{}
        \label{fig:S3-d}
    \end{minipage}
    \caption{Comparing TTA methods for Scenario $3$ on \cifarc{} (a-c) and \pacs{} (d). No TTA method improves accuracy over $\theta_S$ across the settings. The accuracy of SHOT degrades once multiple domains are present. However, in \pacs, SHOT performs better than the other methods and the accuracy grows once $\Delta_T$ contains multiple domains.}
    \label{fig:S3}
\end{figure*}
\myparagraph{Evaluating on $\Delta_{OOD}$.} \figurename~\ref{fig:S2}(c) shows the performance on out-of-distribution (OOD) categories, which are not used for adaptation. The trends resemble those in \figurename~\ref{fig:S2}(a, b). SHOT and NOTE still outperform other methods, and the accuracy steadily increases with increasing $|\mathcal{C}(\Delta_T)|$. The spike in accuracy at $|\mathcal{C}(\Delta_T)| = 8$ is attributed to a similar effect as explained above due to disproportionate prediction to the `Frog' category. Here, due to random sampling, the $\mathcal{C}(\Delta_{OOD})$ contains `Frog' class $4$ times out of $5$ runs.

\myparagraph{Generalization of SHOT in limited categories.} Remarkably, SHOT achieves a better accuracy gain for the test set $\Delta_{OOD}$ (\ref{fig:S2}(b)) compared to $\Delta_T$ (\ref{fig:S2}(c)): with $|\mathcal{C}(\Delta_T)| = 2$ it achieves $1.3\times$ accuracy gain over $\theta_S$ when testing on $\Delta_T$, whereas it achieves $2.2\times$ improvement when evaluating on $\Delta_{OOD}$. This is attributed to the fact that the clustering-based 
pseudo-labeling in SHOT can fix the structure of the embedding vectors from $\mathcal{C}(\Delta_{OOD})$ as well.

\section{Observations on Scenario $\mathbf{3}$: Diverse Distribution Shifts in $\Delta_T$}
\label{sec:scenario3}
Here, we evaluate the TTA algorithms w.r.t. the Scenario $3$ defined in Section~\ref{sec:S3-defn}. \figurename~\ref{fig:S3} analyses the setting in both the \cifarc{} and the \pacs{} datasets.

Here $k$ domains are uniformly randomly chosen from $\mathcal{S}(D_T)$. Further, $m$ samples from each chosen domain are included in $\Delta_T$. For \cifarc{} and \pacs, we choose $m$ as $960$ and $124$, respectively. Aligned with the definition in Section~\ref{sec:S3-defn}, BoTTA evaluates three testing sets.

\myparagraph{Evaluating on $D_T$, $\Delta_T$, and $\Delta_{OOD}$.}
All three testing sets, $D_T$, $\Delta_T$, and $\Delta_{OOD}$ produce consistent observations in \figurename~\ref{fig:S3}(a, b, and c) respectively. The accuracy of $\theta_S$ varies when evaluating $\Delta_T$ and $\Delta_{OOD}$ according to the particular (random) choices of the domains in the experiments. This highlights that domains suffer from the heterogeneous accuracy drop.

Across the three evaluation settings, none of the TTA methods compared here improve the model accuracy significantly over the source model, $\theta_S$. The accuracy of SHOT increases when $|\mathcal{S}(\Delta_T)| \le 3$ and then it degrades significantly at $|\mathcal{S}(\Delta_T)|=4$ before rising again for $|\mathcal{S}(\Delta_T)|\ge 5$. This is attributed to the exact random choices made at different numbers of domains. 

\myparagraph{Generalization of SHOT in multi-domain.}
Notably, the accuracy of SHOT always stays below the source model accuracy, especially when evaluating on $D_T$ and $\Delta_{OOD}$. This means that SHOT hurts a model in the multi-domain scenario and is contrary to the fact that it achieved the best generalization capability with a limited number of categories in $\Delta_T$ in Section~\ref{sec:scenario2}. This happens as the samples from multiple domains are mixed in the embedding space, and its clustering step assigns wrong pseudo labels to the samples across multiple domains.

\myparagraph{Results on \pacs.}
The result on \pacs, shown in \figurename~\ref{fig:S3}(d), tests the model on the adaptation data itself, $\Delta_T$. SHOT has the best accuracy when it is adapted on $3$ domains, which aligns with the results in \cifarc{} in \figurename~\ref{fig:S3}(a) for a smaller number of domains in $\Delta_T$ (up to $3$).

\section{Observations on Scenario 4: Overlapping Distribution Shifts in $\Delta_T$}
\label{sec:scenario4}

Here, we evaluate the Scenario $4$ defined in Section~\ref{sec:S4-defn}. As the corruption types to be overlapped on a sample need to be controlled, this can only be done in \cifarc{} where the corruptions are applied with synthetic functions, i.e., the transformation function $\chi(.)$ has an analytic form for a given corruption type and severity. Thus, we cannot test this in a natural domain-shifted dataset like PACS.

We handpicked the following five pairs of corruption types that may occur together in practice: (Brightness and Zoom blur), (Fog and Impulse noise), (Snow and Defocus blur), (Snow and Motion blur), and (Zoom blur and Gaussian noise). Further, we selected five triplets of corruption types that may occur together, for example, Frost, Fog, and Snow. We use all corruptions with severity of $5$.

\begin{figure}[]
    \centering
    \begin{minipage}{\linewidth}
        \centering
        \includegraphics[width=0.7\linewidth,keepaspectratio]{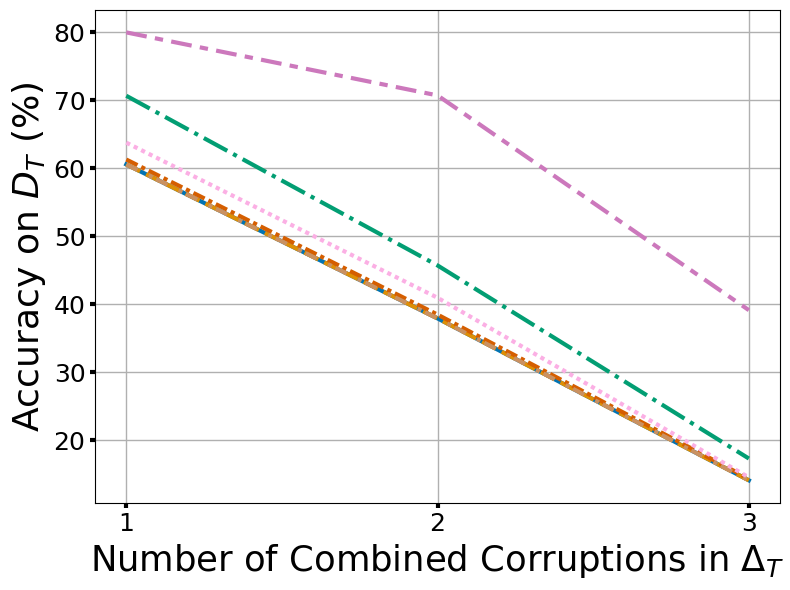}
    \end{minipage}

    \vspace{1mm}
    
    \begin{minipage}{\linewidth}
        \centering
        \includegraphics[width=0.6\linewidth,keepaspectratio,trim={4cm 0 0 0},clip]{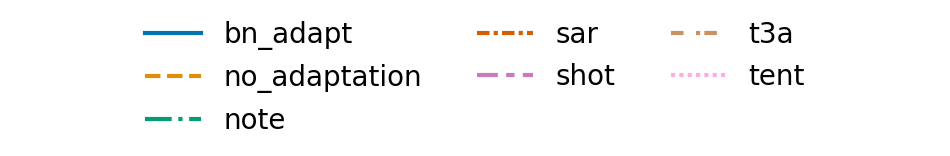}
    \end{minipage}
    \caption{Comparison of TTA methods under test sample corruption by multiple corruption types. SHOT remains the best method in this setting with accuracy $\mathbf{39.1}\%$ when $\mathbf{3}$ types of corruptions are applied at once.}
    \label{fig:S4}
\end{figure}

\figurename~\ref{fig:S4} shows that the accuracy of all the TTA methods reduces as the number of corruptions applied increases. SHOT demonstrates superior robustness to other approaches, maintaining a relatively high accuracy even as the number of corruptions in the mix increases. While all methods experience performance degradation, SHOT achieves an accuracy of $79.9\%$ with singleton domain shifts and declines to $39.1\%$ when three corruptions are applied.

\section{Profiling On-device Resource Consumption}
\label{sec:on_device}
\begin{figure}[!ht]
    \centering
    \begin{minipage}{0.48\columnwidth}
        \centering
        \includegraphics[width=\linewidth,keepaspectratio]{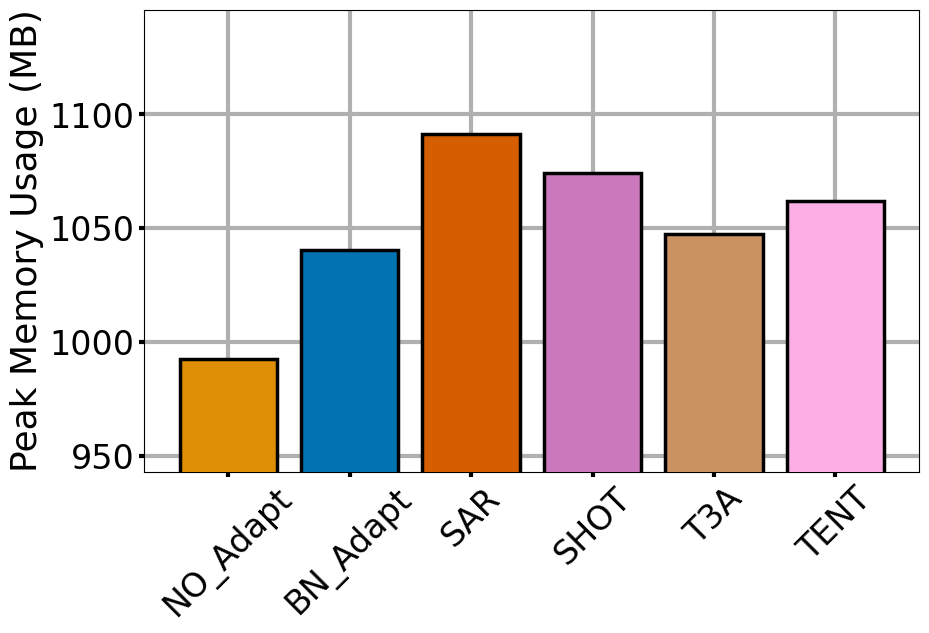}
        \subcaption{}
        \label{fig:S5-a}
    \end{minipage}%
    \hfill
    \begin{minipage}{0.48\columnwidth}
        \centering
        \includegraphics[width=\linewidth,keepaspectratio]{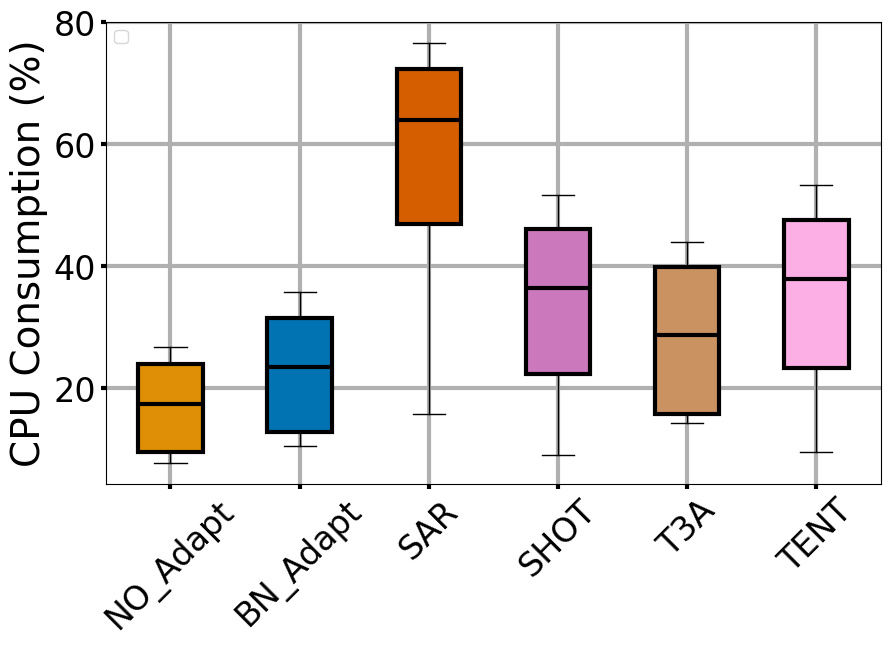}
        \subcaption{}
        \label{fig:S5-b}
    \end{minipage}
    % Row 2
    
    \begin{minipage}{0.48\columnwidth}
        \centering
        \includegraphics[width=\linewidth,keepaspectratio]{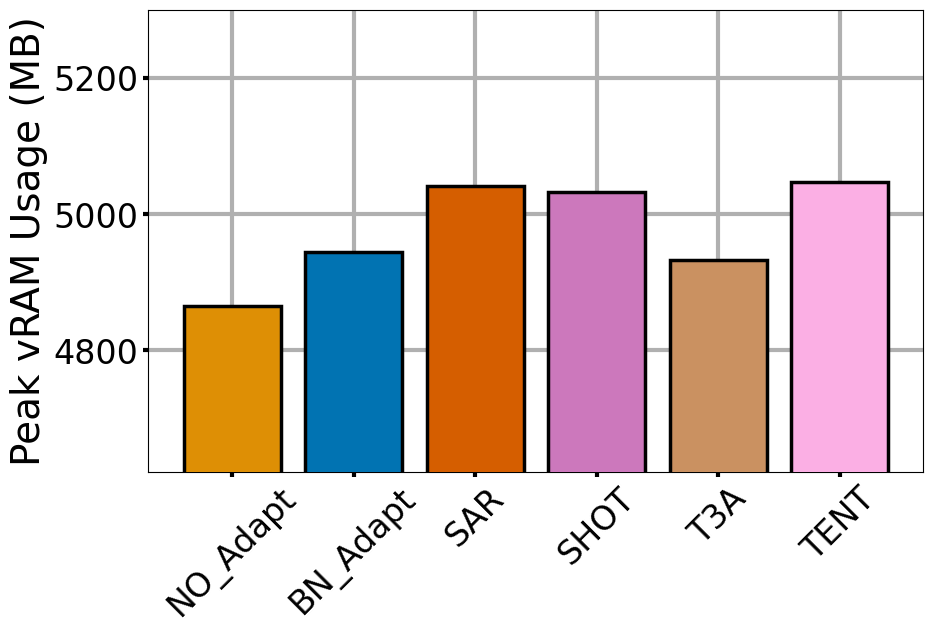}
        \subcaption{}
        \label{fig:S5-c}
    \end{minipage}%
    \hfill
    \begin{minipage}{0.48\columnwidth}
        \centering
        \includegraphics[width=\linewidth,keepaspectratio]{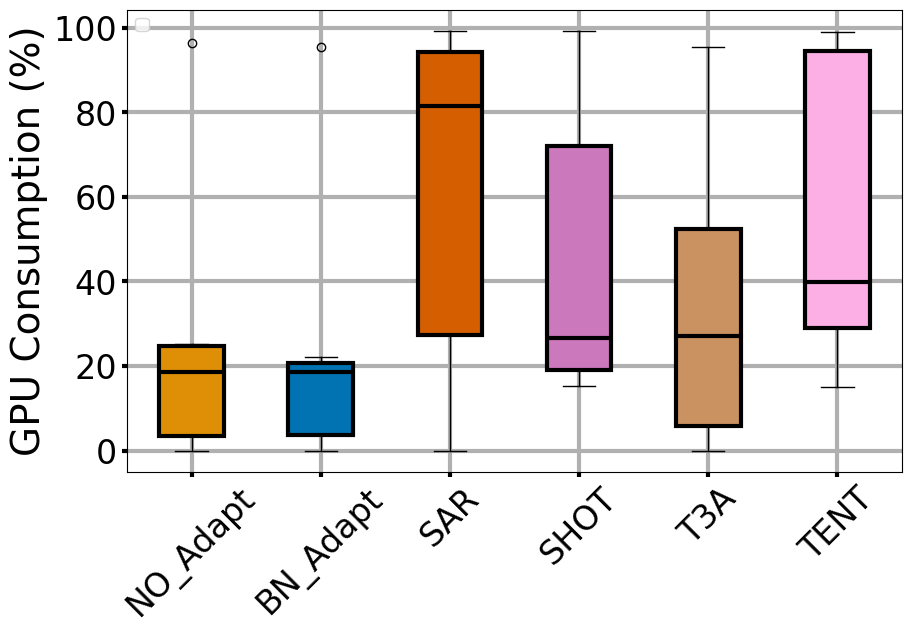}
        \subcaption{}
        \label{fig:S5-d}
    \end{minipage}

    \caption{Resource consumption on (a) and (b) Raspberry Pi-4B ($\mathbf{8}$GB) \& (c) and (d) NVIDIA Jetson Orin. (a) \& (c) report the peak memory consumption. (b) reports the CPU \& (d) reports the GPU consumption.}
    \label{fig:S5}
\end{figure}

Finally, we profile the recent TTA algorithms on a realistic testbed and profile their resource consumption. We first report the peak memory consumption, one of the key deciding factors~\cite{huang2023elastictrainer} to assess the potential of running an ML model on a device. From the results shown in \figurename~\ref{fig:S5}(a) and \figurename~\ref{fig:S5}(c), we observe that most of the TTA algorithms that consider optimization of a proxy test-time objective for fine-tuning the model consume a significant amount of memory. One of the main reasons behind this high memory consumption is the memory intensive gradient computation. Notice that the memory consumption of algorithms like T3A~\cite{iwasawa2021test} is slightly lesser than the other algorithms. This is because T3A does not use any adaptation step and adjusts the trained classifier, the last dense layer of the model, to adapt it at runtime. Interestingly, from \figurename~\ref{fig:S5}(b) and \figurename~\ref{fig:S5}(d), we see a similar pattern for CPU consumption, with SAR consuming the maximum amount of CPU resources. This is due to the overall nature of the optimization approach used by SAR, involving entropy-based filtering and sharpness-aware minimization.

\section{Key Takeaways of BoTTA}
\label{sec:keytake}
In this section, we summarise the key takeaways that we gain from BoTTA.

1. None of the TTA algorithms significantly improve a source model when the adaptation dataset, $\Delta_T$, is small. Across the settings in Scenario $1$, the most accurate method is SHOT. It achieves $1.74\times$ accuracy gain when $|\Delta_T|=256$ whereas it achieves $2.74\times$ gain when using a larger adaptation dataset ($|\Delta_T|=8192$).

2. SHOT and NOTE can significantly improve model generalization when adapting to a partial set of categories. Across most settings, SHOT produces the most accurate models. It produced $2.2\times$ accuracy gain when $\Delta_T$ contains samples from only two out of ten categories in \cifarc{}.

3. None of the TTA methods improve model accuracies when $\Delta_T$ contains diverse distribution shifts. Notably, SHOT hurts the model's performance, especially when the number of domains is high.

4. When multiple distribution shifts overlap in a data sample, all TTA algorithms fail. The accuracy of the best-performing method, SHOT, drops to $39.1\%$ from $79.9\%$ when three corruptions overlap instead of a singleton corruption.

5. Through the experiment on the real-life testbed, we observed that most of the TTA algorithms consume a significant amount of memory. Notably, in the experiments on Raspberry Pi, T3A consumes lower($1.05 \times$) peak memory consumption. In contrast, the most accurate TTA method across many settings, SHOT, consumes higher ($1.08 \times$) peak memory compared to the base strategy of `no adaptation.'

\section{Conclusions}
\label{sec:conclusions}
On-device model adaptation using the in-the-wild target data is important for keeping the model accurate at edge devices. Towards this, recently developed Test Time Adaptation algorithms show initial promise. BoTTA benchmarks a representative set of state-of-the-art TTA algorithms using a novel set of scenarios that a typical edge application will encounter in practice. The scenarios cover data and resource constraints ranging from limited on-device adaptation data to overlapping distribution shifts and system resource constraints. Our exhaustive evaluation suit across two datasets and three model architectures found that though SHOT is the most accurate TTA method across most cases, it fails when adapting to a diverse set of distribution shifts and suffers from high memory consumption while tested on real devices. This shows that more research is necessary to design TTA algorithms suited to edge devices, where we believe our scenarios can be used as a standardized evaluation suit.

% \balance
\bibliographystyle{ACM-Reference-Format}
\bibliography{ref}

\end{document}